\documentclass[10pt,twocolumn,letterpaper]{article}

\usepackage[pagenumbers]{cvpr} 

\usepackage[dvipsnames]{xcolor}

\definecolor{cvprblue}{rgb}{0.21,0.49,0.74}
\usepackage[pagebackref,breaklinks,colorlinks,citecolor=cvprblue]{hyperref}

\usepackage{graphicx}
\usepackage{booktabs}
\usepackage{xspace}
\usepackage{wrapfig}
\usepackage{adjustbox}
\usepackage{enumitem}

\title{Solving Vision Tasks with Simple Photoreceptors Instead of Cameras}

\author{
Andrei Atanov\footnotemark[1]~~$^1$
\;\;
Jiawei Fu\footnotemark[1]~~$^1$
\;\;
Rishubh Singh\footnotemark[1]~~$^1$
\;\;
Isabella Yu~$^{1, 2}$
\;\;
Andrew Spielberg~$^3$
\;\;
Amir Zamir~$^1$ \vspace{2mm}\\
$^1$~Swiss Federal Institute of Technology Lausanne (EPFL) \\
$^2$~Massachusetts Institute of Technology \;\; $^3$~Harvard University \vspace{2mm} \\
\href{https://visual-morphology.epfl.ch/}{https://visual-morphology.epfl.ch/}
}

\definecolor{red}{HTML}{FF0000}

\newcommand{\R}{\mathbb{R}}
\newcommand{\roll}{\mathrm{roll}}
\newcommand{\pitch}{\mathrm{pitch}}
\newcommand{\yaw}{\mathrm{yaw}}
\newcommand{\fov}{\mathrm{fov}}
\newcommand{\bundlesize}{B}
\newcommand{\E}{\mathbb{E}}

\newcommand{\app}{\textcolor{blue}{Supplementary Material}\xspace}
\newcommand{\prab}{PR\xspace}
\newcommand{\pr}{photoreceptor}
\newcommand{\prs}{photoreceptors}

\newcommand{\cprs}{Photoreceptors}

\definecolor{seabornblue}{HTML}{4878D0}
\definecolor{seabornviolet}{HTML}{956CB4}
\definecolor{seaborngreen}{HTML}{6ACC64}
\definecolor{seabornorange}{HTML}{EE854A}
\definecolor{seabornred}{HTML}{D65F5F}

\renewcommand{\emph}[1]{\textbf{#1}}

\begin{document}


\maketitle

\begin{abstract}

A \textit{de facto} standard in solving computer vision problems is to use \textbf{a common high-resolution camera} and choose its placement on an agent 
(i.e., position and orientation) based on human intuition. On the other hand, \textbf{extremely simple and well-designed visual sensors} found throughout 
nature allow many organisms to perform diverse, complex behaviors. In this work, motivated by these examples, we raise the following questions:
\begin{enumerate}
  \item \textbf{How effective} simple visual sensors are \textbf{in solving vision tasks?}
  \item \textbf{What role does their design play in their effectiveness?}
\end{enumerate}
We explore simple sensors with \textbf{resolutions as low as one-by-one pixel}, representing a single photoreceptor. First, we demonstrate that \textbf{just a 
few photoreceptors can be enough to solve many tasks}, such as visual navigation and continuous control, reasonably well, with performance comparable 
to that of a high-resolution camera. Second, we show that the \textbf{design of these simple visual sensors plays a crucial role} in their ability to provide 
useful information and successfully solve these tasks. To find a well-performing design, we present a \textbf{computational design} optimization algorithm and 
evaluate its effectiveness across different tasks and domains, showing promising results. Finally, we perform a human survey to evaluate the 
effectiveness of intuitive designs devised manually by humans, showing that the computationally found design is among the best designs in most cases.

\end{abstract}
\vspace{-1.8em}

\section{Introduction}
Visual sensors provide necessary information about the surrounding world to enable visual perception and problem-solving.
A wide variety of visual sensors are found throughout nature ~\cite{land_animal_2012,banks_why_2015,cronin_visual_2014}, ranging from complex, lens-based eyes that perceive fine-grained signals to extremely simple ones, consisting of only a few photoreceptors that simply capture unfocused light from many directions to create a low-dimensional signal.

\begin{figure}
    \centering
    \includegraphics[width=\linewidth]{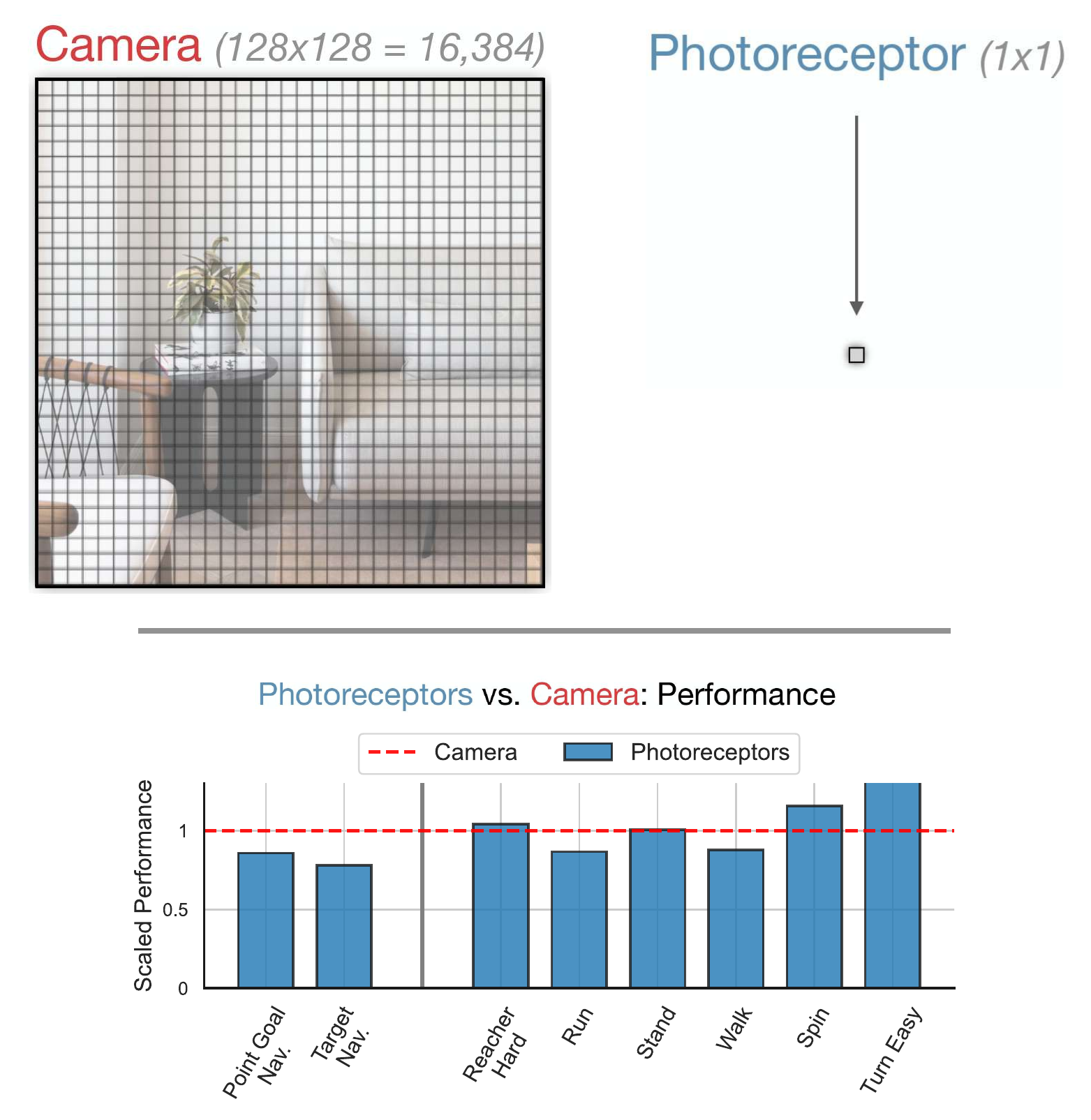}
    \caption{
    \footnotesize
    \textbf{Extremely simple photoreceptor sensors can solve vision tasks reasonably well, comparable to a high-resolution camera.}
    \textit{Top:} We use photoreceptor sensors with a resolution as low as $1 \times 1$, whose dimensionality is 16384 times lower than that of a $128\times 128$ camera sensor (for visualization purposes, the displayed grid in the figure underestimates this factor).
    \textit{Bottom:} We find that even a handful of well-placed photoreceptors can provide sufficient information to solve some vision tasks with reasonably good performance - significantly higher than a blind agent and similar to a more complex camera sensor.
    Our evaluation suite consists of eight vision-based active tasks, including visual navigation using scans of real buildings from the MatterPort3D dataset~\cite{chang_matterport3d_2017} and continuous control tasks from the DeepMind Control suite~\cite{tassa_deepmind_2018}.
    }
    \label{fig:pull}
\end{figure}

In addition to the sensor's type, its position and orientation are also important to its effectiveness.
Strategic placement of even the simplest sensors can enable complex behaviors such as obstacle avoidance, detection of coarse landmarks, and even some forms of predator avoidance~\cite{land_animal_2012}. 
For example, in dragonflies, an upward-facing visual acute zone, i.e., an area with high photoreceptor density, is hypothesized to allow more efficient prey detection by positioning it against the sky instead of a cluttered foliage background. 
Similarly, some species of surface-feeding fish have eyes with horizontal acute zones that allow them to see prey both above and below the water even while entirely submerged by taking advantage of the refractive index of water through their positioning.
This wide variety of designs is believed to emerge as evolutionary adaptations to an animal's specific morphology and the ecology in which it lives~\cite{land_animal_2012}.

In computer vision, on the other hand, the design of visual sensors is mostly represented by one side of the spectrum, namely, complex camera sensors.
Moreover, most effort is spent on algorithmic improvements, leaving sensor design to human intuition.

This paper explores the other side of the spectrum and employs extremely simple visual sensors. 
In particular, we choose visual sensors with a resolution as low as one pixel, representing a single photoreceptor.
One can intuitively think of this as a camera with a 1$\times$1 resolution. 
\textit{We demonstrate that just a few well-designed, i.e., strategically placed and oriented, photoreceptors provide sufficient information to solve many active vision tasks} such as visual navigation, continuous control, and locomotion with a performance much higher than that of a blind agent and close a complex camera sensor (see~\cref{fig:pull}).

Similar to findings in nature, when using simple photoreceptors, we find that designing the sensors' placement, orientation, and field of view (FoV) is essential in achieving optimal performance, and an uninformed (random) design can result in a performance close to that of a blind agent without access to any visual signal.
To find well-performing designs, \textit{we present a computational design optimization method} that optimizes sensors' design for a given agent, environment, and task at hand.
We demonstrate promising results of its effectiveness in improving initial random designs in a variety of domains and allowing us to achieve performance similar to that of the camera sensor.
Finally, to estimate whether humans can devise performant designs, \textit{we conduct a human survey to collect human intuitive designs} and find that the computational design is among the best designs in most cases.


\section{Related Work}


\textbf{Camera Design Optimization.}
This line of work aims at optimizing camera parameters such as lens configuration~\cite{sun_end--end_2021,sitzmann_end--end_2018,ikoma_depth_2021,baek_single-shot_2021,chang_deep_2019}, camera placement~\cite{sanket_morpheyes_2020, krause_efficient_2008,hou_optimizing_2023} and other~\cite{wu_phasecam3d_2019,olague_optimal_2002,vargas_time-multiplexed_2021,tseng_differentiable_2021} to improve downstream performance.
Most of these approaches consider static downstream tasks such as image restoration or depth estimation with a differentiable loss function, which, combined with a differentiable renderer~\cite{wang__2022,tseng_differentiable_2021,sitzmann_end--end_2018}, enable using gradient optimization methods for design optimization.
In this work, we focus on active vision tasks, where the downstream performance is defined by a non-differentiable reward function.
Most similar, \cite{chen_learning_2022} learns an active camera that rotates during the episode but has a design space limited to turning along a single axis.
In contrast to these works, we co-learn both the active vision task and the design of extremely simple photoreceptor visual sensors.

\textbf{Alternative Visual Sensors.}
In addition to RGB camera sensors, prior work in robotics and visual sensing has also made use of time-of-flight sensors~\cite{lange_seeing_2000}, LIDARs and event cameras. These sensors usually produce high-resolution images and have been used in robust 3D mapping and navigation~\cite{may_three-dimensional_2009, prusak_pose_2008, francis_tof-camera_2015, liu_efficient_2021} and obstacle avoidance at high speeds~\cite{falanga_dynamic_2020,xu_taming_2023}. In contrast to these sensors, the simple photoreceptors we explore only provide sparse signals and are much smaller in size compared to other sensors.

\textbf{Computational Design of Robot Morphologies.} 
Since the idea of computationally designing a robot body for a given task is reminiscent of the evolution of organisms, it is not surprising that evolutionary algorithms were prominent early candidates for design, beginning with co-design of form, actuators, and/or controllers \cite{sims1994evolving, hiller2011automatic, cheney2014unshackling, cheney2014evolved}.  
Such methods were even used to computationally design robots built from organic matter \cite{kriegman2020scalable}, including those with the life-like ability to (physically) reproduce \cite{kriegman2021kinematic}.  More efficient co-optimization algorithms emerged \cite{wampler2009optimal, spielberg2017functional, ha2018computational, schaff2022soft, megaro2017designing, spielberg2023advanced, matthews2023efficient} leveraging differentiable simulation \cite{spielberg2023differentiable};
Yet despite their efficiency, these direct optimization methods converge to a single local minimum and are not robust against a wide variety of conditions.  Learning-based approaches have been used to co-design over learned controllers and geometric forms \cite{schaff2019jointly, won2019learning} as well as wholesale shape and topology \cite{zhao2020robogrammar, xu2021multi, yuan2021transform2act, wang2024diffusebot}.  Learning-based approaches for sensing have been sparser, but have natural value in designing agents that are robust against a wide range of environmental stimuli. Sampling-based methods have been used in the design of static infrastructure \cite{schaff2017jointly} and soft robots \cite{spielberg2021co}, but 
to date, the role of vision-based sensing remains mostly unexplored in robot design.

\textbf{ML-Based Discovery.}
Recently, in many fields, machine learning-based search methods proved useful for discovering new optimal designs, e.g., novel drugs~\cite{jumper_highly_2021,popova_deep_2018},  catalysts~\cite{zitnick_introduction_2020,zitnick_spherical_2022} or ~\cite{szymanski_autonomous_2023}.
These methods usually rely on and benefit from large amounts of data to train a model of the underlying process and predict the desired properties of novel designs.
In contrast, in our case of designing visual sensors, there is no such dataset readily available.
We, therefore, rely on exploring the design space using simulation to provide us with the performance of different designs.

\begin{figure*}
    \centering
    \includegraphics[width=\textwidth]{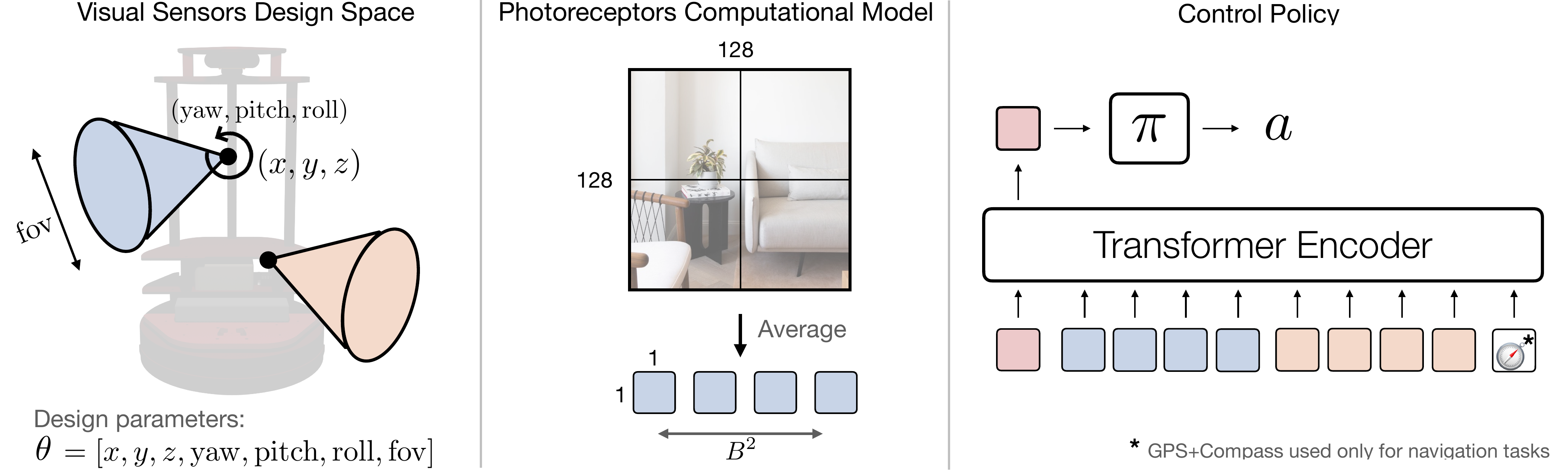}
    \vspace{-10pt}
    \caption{
    \footnotesize
    \textbf{Simple photoreceptor sensor for active vision tasks.}
    \textit{Left:} The design space of visual sensors (\prab or camera).
    We vary the extrinsic (position and orientation) and intrinsic (field of view) parameters for each sensor (either a single \prab or a $B\times B$ grid with shared extrinsic parameters).
    We constrain the position of a sensor to the agent's body.
    \textit{Center:} To implement the \prab{} sensor computationally in common simulators, we render a camera view (e.g., using a pinhole camera model) with the corresponding design parameters and average the signal spatially.
    For a grid sensor, we split an image into equal patches and average each of them spatially to get readings for the corresponding $B^2$ \prab{} sensors.
    \textit{Right:}
    Finally, we pass observations from all sensors (along with \texttt{GPS+Compass} for navigation tasks) through a Transformer encoder to predict the action $a$ that optimizes a task-specific reward function.
    }
    \label{fig:method}
\end{figure*}

\textbf{Zero-Order Optimization}
Zero-order optimization methods aim to optimize an unknown function that can only be evaluated at proposed points and has no other available information, such as gradients~\cite{mockus_bayesian_1975,snoek_practical_2012,hansen_completely_2001,hansen_cma_2006,williams_simple_1992}.
Our design optimization problem for visual sensors, similar to other design optimization problems, falls into this category as no gradients of performance w.r.t. the design are available.
Most similar to our work is \cite{yuan_transform2act_2022}, which also uses a joint design-control optimization method, applying it to the robot's morphology design.
In contrast, we apply it to the problem of designing visual sensors and consider more challenging tasks of visual navigation using scans of real-world buildings.

\section{The Photoreceptor Sensor: Computational Model and Design Space}

This section defines a \pr{} sensor~(\prab), an example of a simple visual sensor explored in this work.
First, we present the employed computational model of a \prab.
Then, we describe the considered design space of these visual sensors.
Finally, since no prior work has explored the usage and design of these sensors, we describe three types of design types explored in this work: random, computational via design optimization, and intuitive.


\subsection{Computational Model of a Photoreceptor}
We define a photoreceptor (PR) as a visual sensor located at a specific point in space that integrates all incoming light from a specific field of view.
Unlike complex, high-resolution lens-based sensors, a \prab{} does not produce a high-resolution image but only provides a low-dimensional signal of the average light intensity (for each color channel).
An analogy from nature is a single photoreceptor placed into a pigment tube, allowing light only from a given direction and field of view. 
In practice, such a sensor can be realized with a photodiode~\cite{emmons_avalanchephotodiode_1967} by restricting its field of view using a casing that prevents it from receiving the light from the entire scene.
The lensless compound eye realized by Kogos et al.~\cite{kogos2020nature} is analogous to a grid of PR sensors introduced below.
Note that our framework, including the design optimization method in \cref{sec:dopt}, is agnostic to this particular choice of a simple visual sensor and can work with any other sensor that can be implemented in a simulator (e.g., see an optimized camera design in \cref{tab:cam-opt-compare}).

Computationally, we implement this definition of the \prab sensor as an averaging of the pixel values of a pinhole camera image.
This approach allows us to model \prab{s} in any simulator that provides a rendering engine without additional implementation costs.
To read the signal $x$ of a \prab with given extrinsic (position and orientation) and intrinsic (the field of view) parameters, we spawn a camera with the same parameters and render its corresponding image view  $I \in \R^{3\times H \times W}$.
Then, we average it's signal along the spatial dimension to get the final value $x^c = \frac{1}{HW} \sum_{i,j} I^{c}_{p},\, x\in \R^3$, where $c \in \{1, 2, 3\}$ each stands for a channel and $p$ for the spatial pixel coordinate.
In addition to a single \prab, we also consider simple sensors using a grid of \prab{s} of size $\bundlesize \times \bundlesize$ for low $B$ ($\leq 8$, in our experiments) that share the same position but have different adjacent fields of view.
We implement such a grid by splitting the image into $\bundlesize^2$ patches and averaging each of them spatially.
See \cref{fig:method}{-Center} for the visualization.
A single grid sensor enables the extraction of useful information (e.g., direction of motion) and makes a useful building block for a simple visual system.
Moreover, using such a grid instead of $\bundlesize^2$ independent photoreceptors results in a significant computational improvement when training in simulation, as it requires rendering only a single image instead of $\bundlesize^2$ images.


\subsection{Design of Visual Sensors}
\label{sec:sensors-design}
\textbf{Design Space.}
We associate each sensor with its 7-dimensional design parameter vector $\theta_i = [\mathbf{x}_i, \mathbf{y}_i, \mathbf{z}_i, \yaw_i, \pitch_i, \roll_i, \fov_i]$, where $(\mathbf{x}_i, \mathbf{y}_i, \mathbf{z}_i) \in \R^3$ is the position in space, which we constrain to be on the agent's body, $(\yaw_i, \pitch_i, \roll_i) \in [0, 2\pi]^2$ is the orientation, and $\fov_i \in [0, 180]$ is the field of view.
See \cref{fig:method}-Left for the visualization.
In our experiments, we use $K \in \{2, 4, 8\}$ sensors and explore sensors represented by a single \prab (a $1\times 1$ grid) and a grid of \prab{s} of sizes $4\times 4$ and $8\times 8$.
This results in a total of $KB^2$ PRs (ranging from 2 to 256 in our experiments) with the visual observation represented as $\{x_{kj}\}_{k,j=1,1}^{K,B^2}$.
We also explore different designs for a camera sensor, in which case we have a single camera sensor and change its vector of parameters $\theta$. 


\textbf{Design Types.}
Choosing a well-performing visual sensor design plays a crucial role in the performance of the final system (e.g., a navigation agent, see \cref{fig:pr-spread}).
In this work, we explore the following three approaches to instantiating the design of a visual sensor:
\begin{itemize}[leftmargin=*,label={}]
    \setlength{\itemsep}{0pt}%
    \setlength{\parskip}{1pt}%
    \setlength{\itemindent}{0.5em} 
    \item \textit{Random design} corresponds to sampling $\theta$ randomly from the design space.
    It sets a baseline for a computational design method, which should result in more performant designs (in cases where they exist in the design space for a given task.)
    \item \textit{Computational design} tailors the sensor parameters for a specific vision task, agent's morphology, and environment, optimizing the corresponding performance of the agent.
    We introduce the employed computational design optimization method in \cref{sec:dopt}.
    \item  \textit{Intuitive design} corresponds to a design devised intuitively by a human.
    Since there is no obvious choice for this design, we perform a human survey, asking participants to devise a design that would lead to the best performance on a given task,  agent and environment.
    We describe the design of the survey in \cref{app:intuitive-survey} and discuss our findings on the effectiveness of human intuition in comparison to computational design in \cref{sec:intuitive}.
\end{itemize}

\section{Simple Photoreceptors are Effective Visual Sensors}

In this section, we demonstrate that simple \pr{s} can be effective visual sensors for solving different active vision tasks.
Specifically, we show that in most cases, an agent equipped with (well-designed) \prab{} sensors significantly outperforms a blind agent without access to any visual signal and achieves performance close to that of an agent with a high-resolution camera sensor.

\subsection{Experimental Setting}
\label{sec:exp-setting}
We perform our experiments with the following active vision tasks.
First, we consider two visual navigation tasks using the Habitat~\cite{szot_habitat_2021} simulator with 3D scans of real apartments from the Matterport3D~\cite{chang_matterport3d_2017} dataset.
Our second set of tasks are continuous control tasks from the DeepMind Control (DMC) Suite~\cite{tassa_dm_control_2020}, which we attempt to solve solely from the vision signal.
Below, we provide a brief description of the experimental setting.
For more detailed information, please refer to~\app.

\textbf{Reinforcement learning background.} 
We consider solving the active visual tasks as the decision-making processes using reinforcement learning in partially observable Markov decision processes (POMDP).
At a state $s_t$, the agent receives an observation $o_t$ which cannot precisely determine the underlying state $s_t$.
Then, the agent applies an action $a_t$, transits to the next state $s_{t+1}$, and receives a reward $r_t$.
Let $\tau$ be the trajectory rollout provided to the agent by iterating the steps above, i.e., $\tau_t=(o_t,a_t,r_t,o_{t+1},\cdots)$.
Assume the agent computes the action $a_t$ with a control policy $\pi$, i.e., $a_t\sim \pi(\cdot|o_t)$.
We find the optimal control policy $\pi^\star$ by optimizing the expected return $\mathbb{E}_{\tau_t\sim\pi}[R(\tau_t)]$, where the return is defined as $R(\tau_t)=\Sigma_{i=0}\gamma^ir_{t+i}$ and $\gamma$ is the discount factor. In our experiments, we use Proximal Policy Optimization (PPO)~\cite{schulman_proximal_2017} to optimize the control policy $\pi$.

\textbf{Navigation in Habitat.} We train navigation agents for \texttt{PointGoalNav} and \texttt{TargetNav} tasks in 3D replicas of real houses from the Matterport3D dataset~\cite{chang_matterport3d_2017} using the Habitat simulator ~\cite{szot_habitat_2021}. 
In \texttt{PointGoalNav}, the agent is randomly spawned in an environment and needs to navigate to a target coordinate.
The agent observes an egocentric \texttt{RGB} view and its current position (coordinate) and orientation through the \texttt{GPS+Compass} sensor. We measure the performance of an agent using `SPL' (Success weighted by Path Length)~\cite{anderson_evaluation_2018}, which quantifies the performance relative to the optimal trajectory.
In \texttt{TargetNav}, the agent is also equipped with the egocentric \texttt{RGB} sensor and the \texttt{GPS+Compass} sensor. 
In contrast to \texttt{PointGoalNav}, the agent does not receive a target coordinate but is asked to navigate to a green sphere that is randomly placed in the house at a height of 1.5m from the floor. This task, therefore, requires exploration and target identification by design. We therefore measure the performance of an agent using the success rate; that is, whether or not it finds the target sphere.

\textbf{Continuous Control in DMC.} We train continuous control agents using the MuJoCo simulator~\cite{todorov_mujoco_2012} on the DeepMind Control (DMC) benchmark~\cite{tassa_dm_control_2020}.
DMC provides a variety of continuous control tasks, including reaching, manipulation, locomotion, etc.
In our context, we focus on learning the control policy that receives only visual information from either photoreceptors or a camera.
We consider using the following six tasks: \texttt{Reacher:Hard}, \texttt{Walker:Stand}, \texttt{Walker:Walk}, \texttt{Walker:Run}, \texttt{Finger:Spin}, and \texttt{Finger:Turn Easy} (see \cref{app:dmc-setting} and the original DMC video\footnote{\url{https://www.youtube.com/watch?v=rAai4QzcYbs}} for a more detailed description of tasks.)

\begin{figure}
    \centering
    \includegraphics[width=\linewidth]{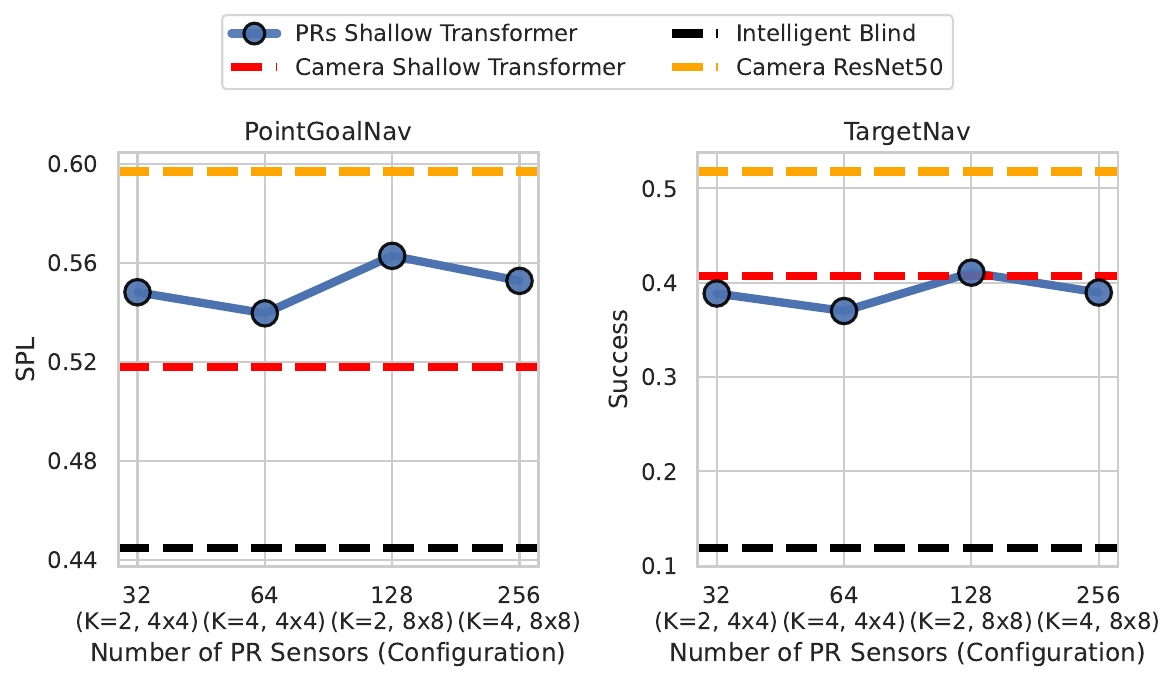}
    \caption{
  \textbf{Photoreceptors are effective visual sensors for navigation tasks.}
  We compare the performance of agents trained with different visual sensors - a varied number of photoreceptors, a camera, or no visual sensor (intelligent blind) - on visual navigation tasks.
  When scaling the number of PRs, we use configurations of $K\in\{2,4\}$ grids of sizes 4$\times$4 and 8$\times$8.
  In all cases, we report the best design found by our design optimization method (see~\cref{sec:dopt}), including the camera design.
  For the camera baseline, we report performance when using the same shallow 3-layer Transformer encoder as for PRs and the ResNet-50 backbone, a default choice in the literature (``gold standard''), for a fair comparison.
  \textit{Even with a handful of photoreceptors, PR agents significantly outperform blind agents and achieve performance closer to or better than that of the camera agent with the same shallow encoder} (getting close to the gold standard.)
  }
    \label{fig:cam-vs-pr-nav}
\end{figure}

\textbf{Architecture.}
\cref{fig:method}-right illustrates our control policy architecture.
We model the policy $\pi_w(a_{t+1}\,|\,o_t)$ using a simple three-layer Transformer~\cite{vaswani_attention_2023} backbone that encodes the current observation $o_t$ from a visual sensor, camera or \prab{s}, and \texttt{GPS+Compass} for navigation, and a small MLP that predicts the next action $a_{t+1}$.
For the \prab-based policy, we use $\{[p_{kj}, \theta_k, e_j]\}_{kj}$ as input tokens, where $\theta_k$ is the design vector of the $k^\text{th}$ grid sensor, and $e_j$ is the trainable positional embedding of the $j^\text{th}$ \prab in the grid.
For the camera-based policy, similar to ViT \cite{dosovitskiy_image_2021}, we split the input image into \texttt{16x16} patches $\{x_{ij}\}$, flatten them, and add positional embeddings and the design vector $\theta$ to construct the final input tokens for the encoder:  $\{[\overline{x}_{ij}, \theta, e_{ij}]\}_{ij}$.
Since we use a single camera, one can omit the design embedding $\theta$, but we keep it for consistency and as we use it in the design optimization method described in \cref{sec:dopt-method}.

\begin{figure}
    \centering
    \begin{subfigure}[b]{.8\linewidth}
        \centering
        \includegraphics[width=\linewidth]{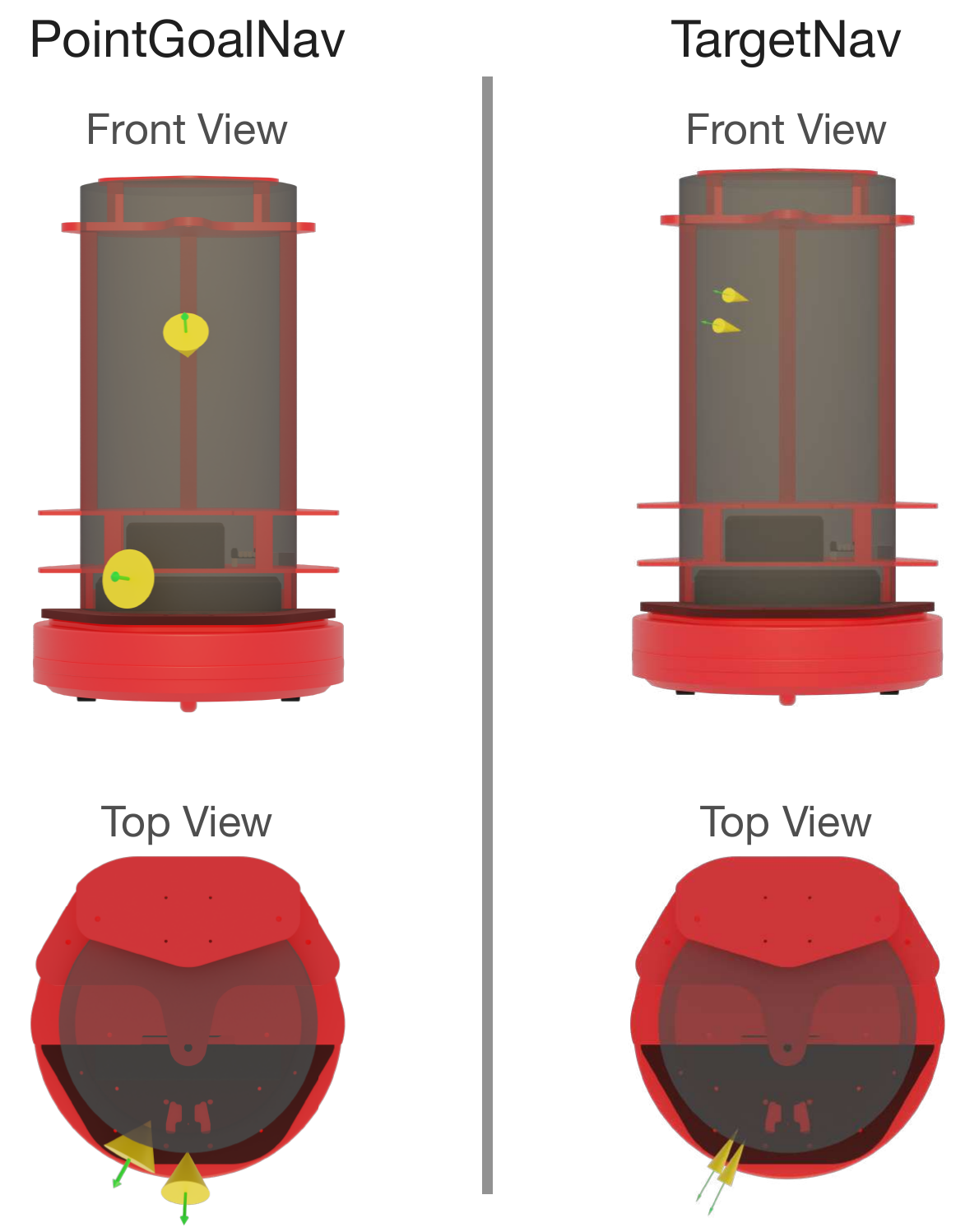}
    \end{subfigure}
    \caption{
  \textbf{Best photoreceptor design visualizations.}
  We visualize the best-performing photoreceptor designs for both \texttt{PointGoalNav} (left) and \texttt{TargetNav} (right) tasks.
  These are computational designs found by the proposed design optimization method (see~\cref{sec:dopt}.) 
  Both designs contain a total of 128 PRs in the configuration of $K=2$ grids of size 8$\times$8.
  While the depicted designs might appear unintuitively irregular, they both result in good performance as seen in \cref{fig:cam-vs-pr-nav} and improve upon the random design initialization (see~\cref{fig:dopt-perf}).
  In addition, \cref{fig:pr-spread} shows that a random, uninformed design statistically does not lead to similar high performance. Therefore, \textit{there is a specific structure to this design}, albeit hard to understand intuitively.
  }
    \label{fig:comp-pr-viz}
\end{figure}

\textbf{Control baselines.}
To estimate the effectiveness of the \prab sensor, we use the following two baselines.
\begin{itemize}[leftmargin=*,label={}]
    \setlength{\itemsep}{0pt}%
    \setlength{\parskip}{1pt}%
    \setlength{\itemindent}{0.5em} 
    \item The \textit{Intelligent blind} agent does not receive any visual signal and shows what performance can be achieved just by utilizing the structure of the problem and environment.
    It does not receive any input in DMC and receives only \texttt{GPS+Compass} in both navigation tasks.
    \item The \textit{Camera} agent receives a high-resolution image signal from a camera sensor.
    This baseline provides a comparison to a \textit{de facto} standard for solving (active) vision tasks.
    For the navigation tasks, we use the resolution of \texttt{128$\times$128} and the found computational design for the camera sensor as we found it to perform better than the default intuitive design from the Habitat simulator in both tasks (see~\cref{tab:cam-opt-compare} and \cref{sec:dopt-exp}).
    For DMC tasks, we choose the best performance between the default 3rd-person view camera with the resolution of \texttt{84 $\times$ 84} (standard choice in the literature~\cite{yarats_mastering_2021,laskin_reinforcement_2020}) and an egocentric camera with an intuitive design (e.g., forward-looking camera on top of the torso for the walker agent), which we also find to perform better in some cases.
    We also use a convolutional architecture similar to~\cite{tassa_deepmind_2018} for a fair comparison, as we find it to perform better.
\end{itemize}

\begin{figure*}
    \centering
  \includegraphics[width=\textwidth]{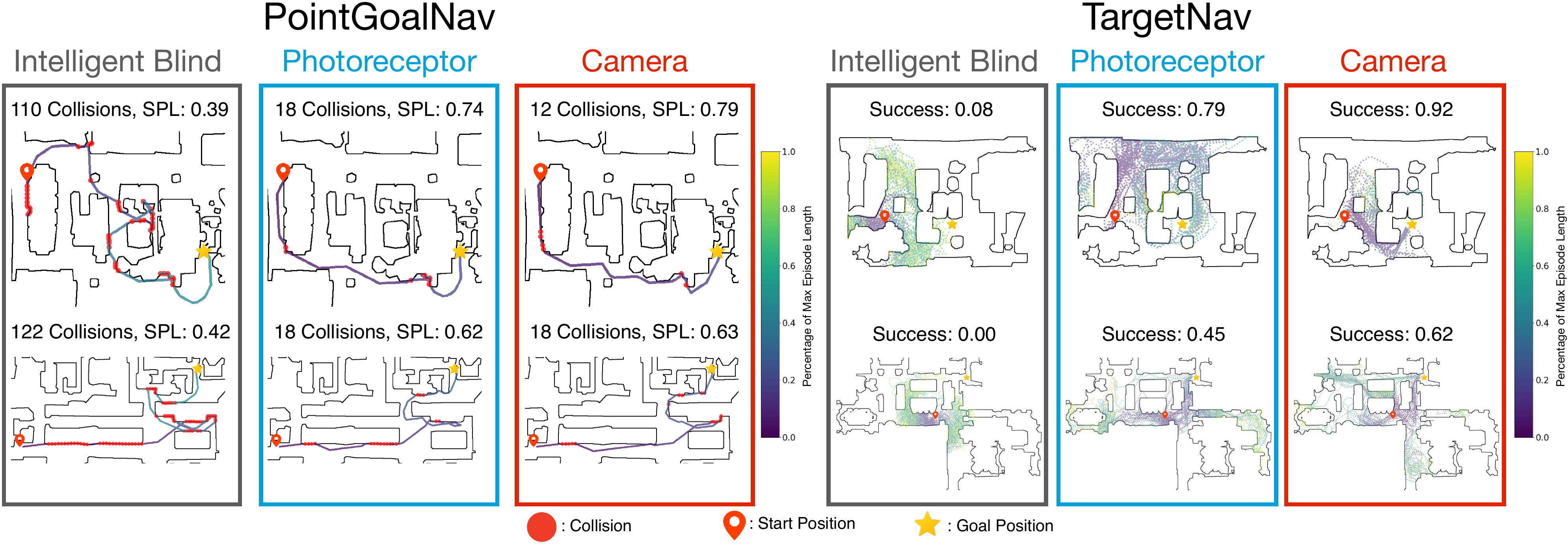}
  \vspace{-20pt}
  \caption{
  \footnotesize
  \textbf{Left: In \texttt{PointGoalNav}, photoreceptors enable collision avoidance and choose efficient trajectories.}
  For each agent, we plot the trajectories from two episodes from unseen test scenes.
  The red dots denote actions that result in a collision. We find that the PR agent can avoid collisions and choose an efficient trajectory similar to that of the camera agent.
  \textbf{Right: In \texttt{TargetNav}, photoreceptors enable efficient exploration and target detection.}
  We plot 50 trajectories for each agent and an unseen test scene.
  PR agents are able to explore novel scenes efficiently (see spread dark points that indicate early steps in the episode compared to the blind agent) and successfully find the target in most cases, approaching the performance of the camera agent. 
  }
  \label{fig:2d-traj-compare}
\end{figure*}

\begin{figure*}
    \centering
    \includegraphics[width=\textwidth]{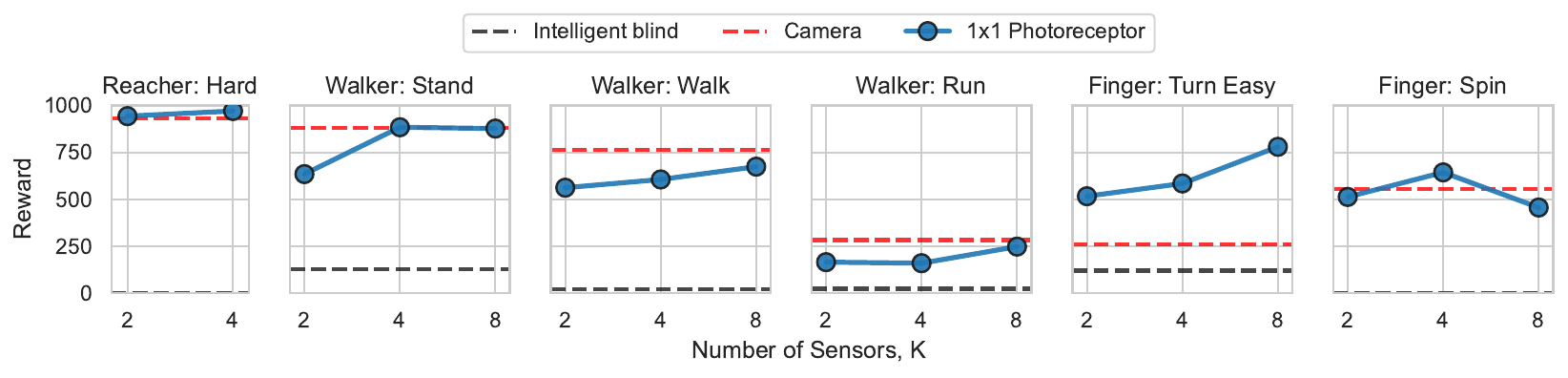}
    \vspace{-20pt}
    \caption{
    \footnotesize
    \textbf{Simple photoreceptor sensors lead to high performance similar to that of the camera sensor in most tasks from the DeepMind Control suite.}
    We show the reward achieved by agents equipped with photoreceptors, a camera, and no visual sensor (intelligent blind).
    For all tasks, the maximum possible reward is 1000.
    We use $K\in\{2,4,8\}$  photoreceptor sensors of resolution 1$\times$1 (except for the \texttt{Reacher:Hard} task where we found four PR sensors already achieving close-to-maximum performance).
    We choose the best-performing design for each PR configuration among random and computational designs (the computational design is the best in most cases; see \cref{fig:dopt-perf} for comparison).
    For the camera sensor, we chose the best-performing design between the third-person view, a \textit{de facto} standard design for these tasks, and an intuitive egocentric camera, which we found to perform better in some cases.
    \textit{In most tasks, we find that just a few \prab{s} significantly outperform the blind agent and perform on par with the camera sensor.}
    }
    \vspace{-1em}
    \label{fig:cam-vs-pr-dmc}
\end{figure*}

\subsection{Photoreceptors Achieve Performance Close to a Camera}
\label{sec:cam-vs-best-pr}

\textbf{Visual Navigation.}
\cref{fig:cam-vs-pr-nav} shows that for both \texttt{PointGoalNav} and \texttt{TargetNav} tasks, even a simple visual sensor consisting of two 4$\times$4 PR grids (32 PRs) provides useful information, allowing the corresponding agent to outperform significantly an intelligent blind agent without a visual signal.
The PR agents match the performance of the camera agent using the same shallow 3-layer Transformer encoder as for PRs, while having a visual signal bandwidth of only $\approx1\%$ of that of the \texttt{128$\times$128} camera sensor.
When compared to the camera agent using the ResNet-50 encoder, a default choice in
the literature (“gold standard”) for a fair comparison, PR agents still perform reasonably well. \cref{fig:comp-pr-viz} shows visualizations of the best-performing photoreceptor designs for both \texttt{PointGoalNav} and \texttt{TargetNav} tasks.


\cref{fig:2d-traj-compare} further demonstrates exemplar trajectories for each type of agent for each task.
In the \texttt{PointGoalNav} task, where the target position is known and the main challenge is to navigate to it efficiently, we find that the PR agent follows more optimal trajectories (as measured by SPL) and effectively uses the visual signal to avoid collisions similar to the camera agent.
In the \texttt{TargetNav} task, on the other hand, the target's position is unknown and can only be identified visually, and it is important to explore a scene efficiently.
\cref{fig:2d-traj-compare}-right shows that both the PR and camera agents explore the scene much more efficiently, achieving a much higher success rate compared to the blind agent.

\begin{figure*}
  \centering
    \includegraphics[width=0.95\textwidth]{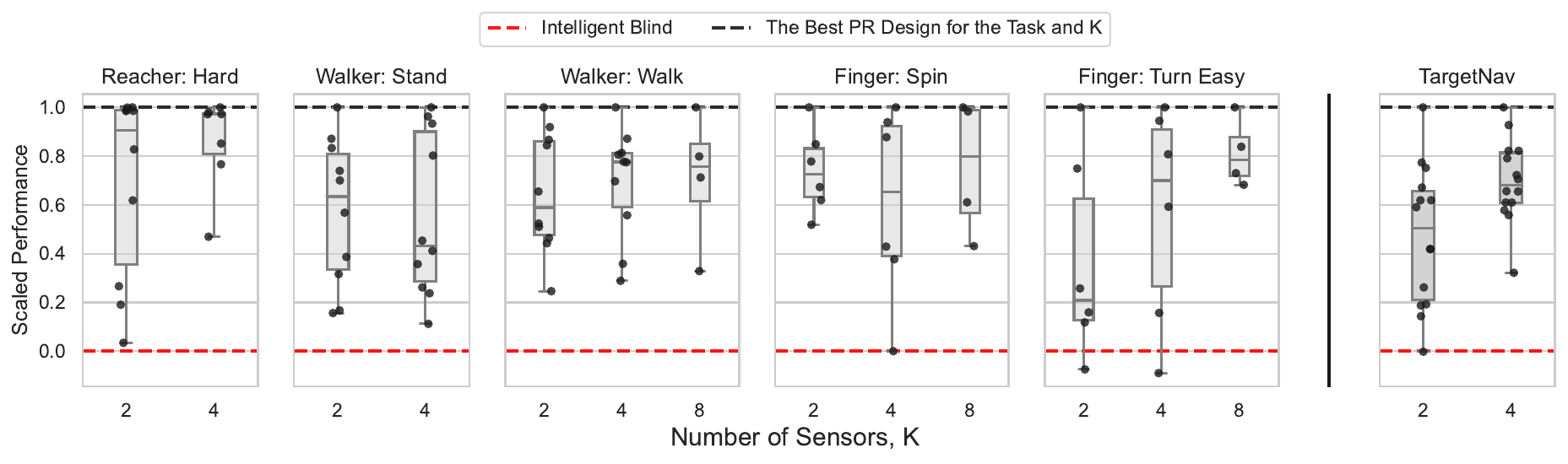}
    \vspace{-10pt}
    \label{fig:pr-spread-dmc}
  \caption{
  \footnotesize
  \textbf{The design of simple photoreceptor sensors plays a crucial role in their effectiveness.}
  We show the performance of different \prab designs for DMC (left) and visual navigation tasks (right), including random, intuitive, and computational designs.
  For each number of photoreceptors $K$ ($K$ grids of size $4\times 4$ for \textit{TargetNav}) and task, we scale the performances of different designs by the performances of the best-performing design (corresponds to 1) and the blind agent (corresponds to 0).
  \textit{We find that a good design is crucial in achieving high performance, and poorly designed sensors can lead to a significant performance drop.}
  }
  \label{fig:pr-spread}
\end{figure*}

\textbf{Continuous Control in DMC.}
\cref{fig:cam-vs-pr-dmc} 
demonstrates that for most of the considered tasks, an agent with a few $1\times$1 photoreceptors significantly outperforms the blind agent and performs closely to the camera agent.
Adding more sensors or increasing the grid sizes further improves the performance in most cases.
However, we find that having a higher-resolution signal leads to a performance drop in some cases.
We conjecture that the main reason is the suboptimal design.
Indeed, as we find in~\cref{sec:dopt}, the design optimization algorithm fails to find an optimal design in some cases.

We note that, in general, the camera agent can achieve higher performance by utilizing, for example, more specialized and tuned reinforcement learning algorithms~\cite{yarats_mastering_2021}, data augmentation techniques~\cite{laskin_reinforcement_2020}, and/or training for more steps~\cite{tassa_deepmind_2018}.
In contrast, we employ a standard PPO~\cite{schulman_proximal_2017} algorithm with a relatively short number of steps compared to~\cite{tassa_deepmind_2018} (5x).
Note, however, that PR agents achieve reasonably good performance even compared to the maximum possible reward of 1000.
In addition, we find that, for example, a longer training of the $K=4, 1\times 1$ PR design leads to an improved reward of 930 compared to the original 605, suggesting that the performance of PR agents can also be improved further by specializing the learning algorithm or longer training.

\section{Visual Sensors Design Optimization}
\label{sec:dopt}
\cprs~can be effective visual sensors, as we show in the previous section.
However, how does one design such a visual sensor?
Where should one place each \prab, and in which direction should they point to provide the most useful information for a given task, environment, and agent's morphology?
In this section, we first show that design choice is essential to achieving good performance.
We then introduce a \textit{computational design optimization} method that optimizes the design for a given agent, task, and environment and shows promising results in improving initial designs across multiple tasks.
Finally, we \textit{perform a human survey} to provide a baseline for an intuitive design, finding that the computational design is among the best designs.

\subsection{Design is Important for the Effectiveness of Photoreceptors}

How does the design of \prs~influence the final performance of a control policy?
\cref{fig:pr-spread} shows that the performance of a poor design can drop drastically compared to the best design for the corresponding task.
We find that some designs result in performance similar to that of a blind agent, suggesting that the visual signal does not provide any useful information.
These results signify the importance of a design optimization algorithm to find good-performing designs automatically.

\subsection{Computational Design via Joint Optimization}
\label{sec:dopt-method}

The design $\theta$ of the visual sensor(s), either \prab{s} or camera, defines what observation the agent receives at each step, i.e., $o_t \triangleq o_t(\theta)$ and what design-specific control policy $\pi_w \triangleq \pi_w^\theta$ with what performance will be learned.
To find the best design $\theta^*$, one would need to find the design that leads to training the best-performing design-specific control policy, resulting in a bi-level optimization problem
$\max_\theta\,\max_w \E_{\tau\sim \pi_w} R(\tau)$.
However, training the design-specific control policy $\pi_w^\theta$ in an inner loop for every new design would make this process prohibitively expensive.

Similar to \cite{yuan_transform2act_2022}, we, instead, cast this problem as \textit{joint optimization} and amortize the costs of training multiple design-specific policies by training a single \textit{``generalist''} policy that implements control for different designs.

\begin{figure}[tb]
  \centering
  \includegraphics[width=\columnwidth]{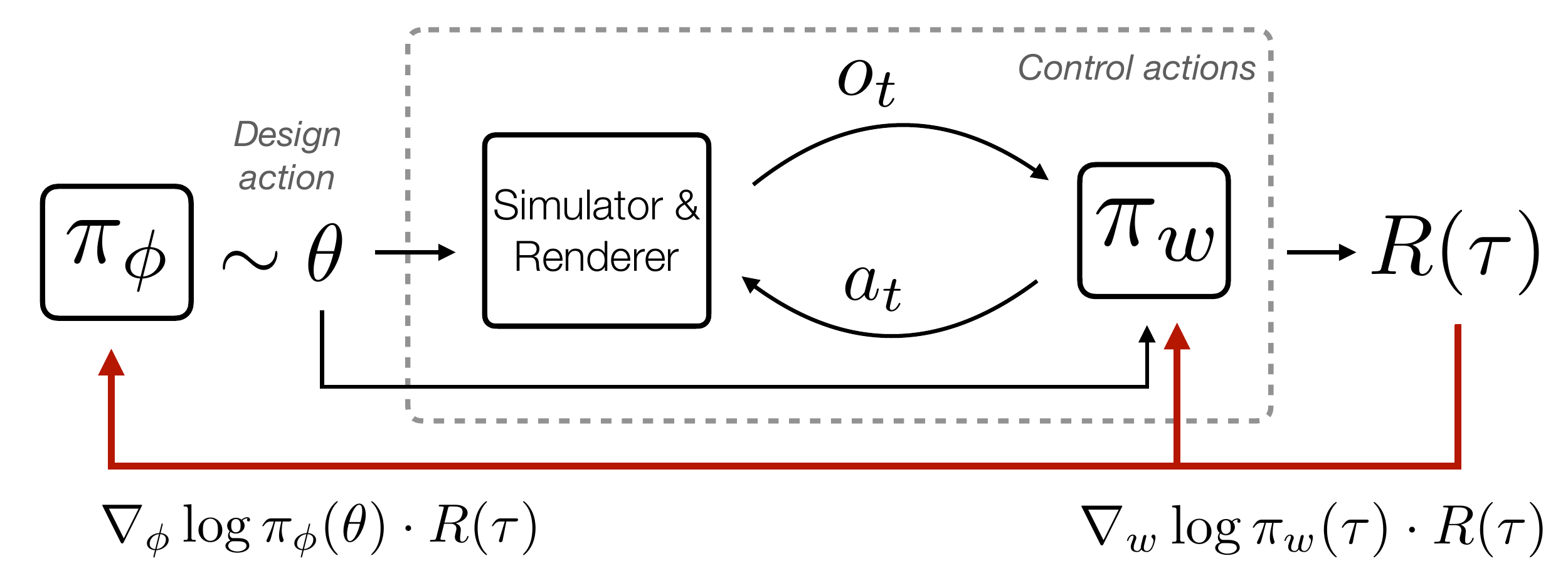}
  \caption{
  \footnotesize
  \textbf{Computational design via joint optimization.}
  We develop a computational design optimization method that jointly trains a control policy $\pi_w$ and a design policy $\pi_\phi$.
  During each episode, first, the design policy samples the design parameters $\theta$ for the visual sensor to be used during the episode, and then the control policy predicts actions based on the current observation governed by the current design.
  Instead of training design-specific control policies for each design, we train a design-conditional policy that implements control for a given design.
  At the end of the episode, we use the final reward to update the design and control policies jointly.
  We use a Gaussian design policy and take its mean as the final design $\theta^*$ after training.
  }
    \label{fig:dopt-method}
\end{figure}

We achieve this by conditioning the policy on the current design vector: $\pi_w(a_{t+1}\,|\,o_t, \theta)$.
In practice, we use the same design-specific architecture described in \cref{sec:exp-setting}, as it already receives the corresponding design vector $\theta$ as part of the input.
To optimize the design, we define a \textit{design policy} $\pi_\phi(\theta)$ and optimize its parameters jointly with the parameters of the control policy:
\begin{equation}
    \label{eq:dopt}
    \phi^*, w^* = \arg\max_{\phi, w}\,\E_{\Tilde{\theta} \sim \pi_\phi(\theta)} \E_{\tau \sim \pi_w(a_t\,|\,o_t, \Tilde{\theta})} \,R(\tau).
\end{equation}
In practice, this implies extending the original decision process with an additional design action step $a_0 \triangleq \theta$ at the beginning of each episode.
We set the corresponding observation and reward to zero, $o_0=\mathbf{0},\,r_0=0$.
One, however, can use a design-dependent reward $r_0$ to favor specific designs, \eg, low-cost ones.
We, then, use the same PPO algorithm and update the control policy using control actions $a_{1:T}$ and the design policy using the design action $a_0$ from each rollout $\tau$.
See \cref{fig:dopt-method} for the visualization, and refer to \cref{app:dopt} for further implementation details.

\textbf{Design Policy.}
We model the design policy as a Gaussian distribution over the design parameters $\pi_\phi(\theta) = \mathcal{N}(\theta \,|\,\mu, \mathrm{diag}(\sigma))$, where $\phi=(\mu, \sigma), \, \mu,\sigma \in \R^{K \times 7}$.
After training, we use $\theta^*=\mu$ as the final optimal design.
Note that while the distribution models each sensor independently, the final design of each sensor is informed of each other by virtue of being optimized together.

\textbf{Generalist Control Policy.}
Training the generalist policy allows, in principle, to amortize the costs of training design-specific policies by ``knowledge'' and parameter sharing.
Modeling and training such a policy that implements control for \textit{all} possible designs can still have high memory and computing costs.
Note, however, that \cref{eq:dopt} only needs the policy that implements control for (likely) samples $\Tilde{\theta}$ from the current design policy to provide it with a \textit{local} direction for the improvement.
Thus, we only need to train a \textit{local generalist} policy and control the locality by the variance of the design policy, which in the limit of low variance allows approximating such a control policy with a linear dependency on design~\cite{mackay_self-tuning_2019,lorraine_stochastic_2018}.
In practice, we initialize the variance $\sigma$ to allow training the local generalist policy that performs similarly to design-specific policies and, thus, provides a good signal for a design update.

\subsection{Design Optimization Experiments}
\begin{figure}[tb]
  \centering
  \includegraphics[width=0.9\columnwidth]{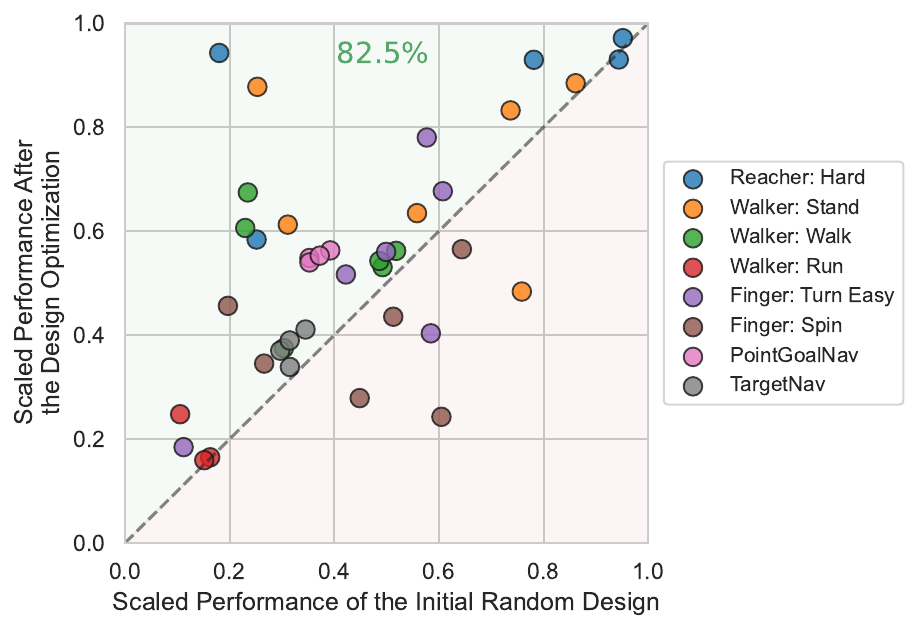}
  \caption{
  \footnotesize
  \textbf{Design optimization improves upon the initial random designs.}
  We report the performance of the random initial design ($x$-axis) and the computational design after the optimization ($y$-axis).
  For DMC tasks, we scale the performance by dividing by the maximum possible reward of 1000.
  The dashed line visualizes the cases where the performance of the design did not change after optimization.
  The green and red regions signify success and failure cases, respectively.
  We find that the proposed optimization method successfully improves the performance of the initial random design in most cases ($82.5\%$ of points are in the green region).
  }
    \label{fig:dopt-perf}
\end{figure}

\label{sec:dopt-exp}
In this section, we demonstrate the effectiveness of the proposed design optimization method.
We show that it can improve the performance of the initial design for both photoreceptors and camera sensors.

\textbf{Design optimization improves the performance} upon the initial design guess.
\cref{fig:dopt-perf} shows that in the majority of cases, the design optimization method improves the initial random design.
For DMC, we run the design optimization from two random initializations for each setting (task and $K$) and find that while it improves their performance, the best-performing computational design depends on the initialization (e.g., reaching the reward of 375 and 562 for two initializations for the \texttt{Walker:Walk} tasks and $K=2$).
This suggests that the optimization landscape might contain multiple local optima, and improving the exploration abilities of the design optimization method is an important research direction for finding the best-performing designs.

\begin{table}[]
    \centering
    {%
\begin{tabular}{@{}lcc@{}}
\toprule
Camera Design & \begin{tabular}[c]{@{}c@{}}PointGoalNav\\ {\footnotesize \textit{(SPL)}}\end{tabular} & \begin{tabular}[c]{@{}c@{}}TargetNav\\  {\footnotesize \textit{(Success Rate)}}\end{tabular} \\ \midrule
Intuitive     &            0.447     &            0.363     \\
Computational &    \textbf{0.518}    &    \textbf{0.405}    \\ 
\midrule
Blind Agent   &            0.445   &              0.119     \\ \bottomrule
\end{tabular}

        }%
    \caption{\textbf{Design optimization finds a better-performing camera design for navigation tasks.} Compared to intuitive design, the computationally found design shows significant improvement in performance on both navigation tasks for a camera agent.}
    \label{tab:cam-opt-compare}
\end{table}


\textbf{Design optimization improves the ``default'' intuitive camera design.}
We also apply the design optimization to explore if we can improve the intuitive camera design used by default in the Habitat AI~\cite{szot_habitat_2021} simulator.
\cref{tab:cam-opt-compare} shows that the agent using a computationally designed camera outperforms the one using the default intuitive design in both navigation tasks.

\subsection{Intuitive Designs}
\label{sec:intuitive}

\begin{figure}
    \centering
    \includegraphics[width=\linewidth]{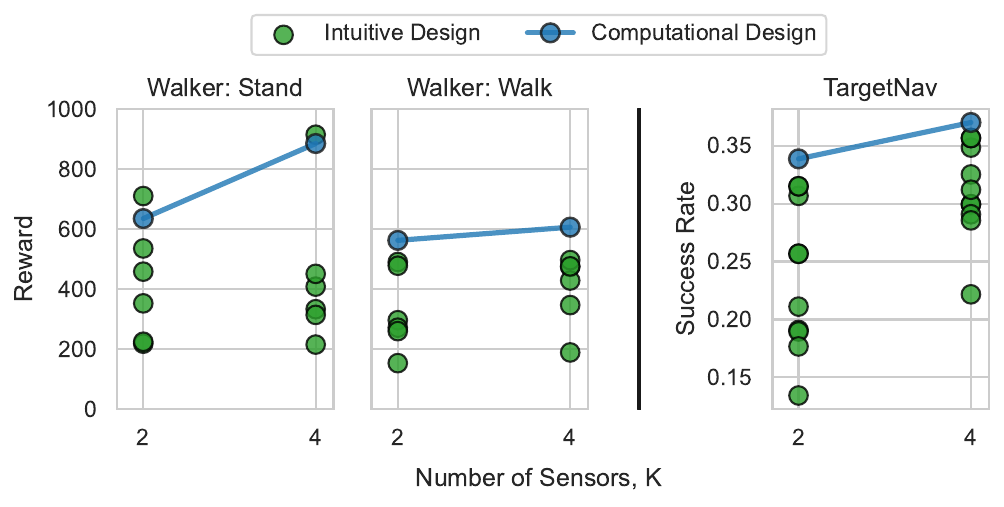}
    \vspace{-2em}
    \caption{
    \footnotesize
    \textbf{Comparison of computational and human intuitive designs.}
    We collect intuitive designs for \texttt{Walker: Stand}, \texttt{Walker: Walk}, and \texttt{TargetNav} tasks via a human survey (see~\cref{sec:intuitive} for details).
    We use the resolution of $1\times 1$ for both DMC tasks and $4\times 4$ for  \texttt{TargetNav}.
    We compare the collected intuitive designs with the best computational design found via design optimization.
    \textit{The computational design is among the best designs in all settings.}
    }
    \label{fig:intuitive}
\end{figure}
In this section, we explore the effectiveness of human intuition in engineering well-performing designs of simple photoreceptor sensors.
Since there is no single obvious way to design such visual sensors, we conducted a human survey to collect intuitive designs.
We ask participants to design the design parameters of visual sensors in our defined design space.
We ask participants to find the \pr{s} design parameters for a given morphology and task.
We collect eight designs for the \texttt{TargetNav} visual navigation task and six designs for the \texttt{Walker} agent in DMC and evaluate them on \texttt{Walk} and \texttt{Stand} tasks.
We provide a more detailed description of the survey setting in \cref{app:intuitive-survey}.

\cref{fig:intuitive} shows that the best human intuition can provide well-performing designs, and computational design is among the best designs (or the best one) in most cases.
We also find a high variance in the performance of different intuitive designs in all settings, signifying the importance of a computational approach to visual design.



\subsection{Do designs transfer between tasks?}

\begin{figure}
    \centering 
    \includegraphics[width=\linewidth]{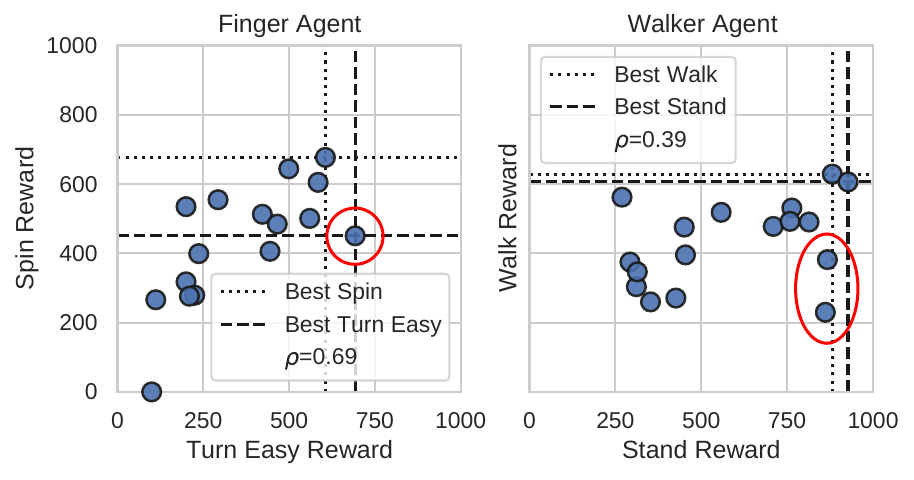}
    \caption{
    \footnotesize \textbf{Do designs transfer across tasks?}
    We consider two pairs of tasks for two agent morphologies from DMC Suite: \texttt{(Turn Easy, Spin)} for \texttt{Finger} (left) and \texttt{(Stand, Walk)} for \texttt{Walker} (right).
    Each point on a plot represents one design for the corresponding agent.
    $x$ and $y$ axes show their performance on the two corresponding tasks.
    $\rho$ is the Spearman's rank correlation.
    In both cases, we find that a correlation exists in general, but some designs might still be primarily tuned for one task (shown in red).
    }
    \label{fig:design-transfer} 
\end{figure}

Optimizing a design for a given agent and task and deploying it can be a time-consuming process.
One would wish to have a visual sensor design that can be optimized and deployed once and recycled for different downstream applications of the same robot without needing to repeat the process.
We compare the performance of different designs we collect in this work (random, intuitive, and optimized) on two pairs of tasks for the same agent morphology.
\cref{fig:design-transfer} shows that a general trend suggests that one can optimize the design for one task and recycle it for another one.
However, there are some designs that can underperform when transferred, especially for the \texttt{Walker} agent.
This means that to find a transferable design during design optimization on one task, some form of regularization needs to be included in addition to the performance only to avoid such cases.

\subsection{Evaluation in the Real World}

To evaluate generalization and ensure that the strong performance of photoreceptors is not confined to simulators, we conducted the target navigation experiment (without access to GPS+Compass sensor) in a real-world setting.

\begin{figure}
    \centering 
    \includegraphics[width=\linewidth]{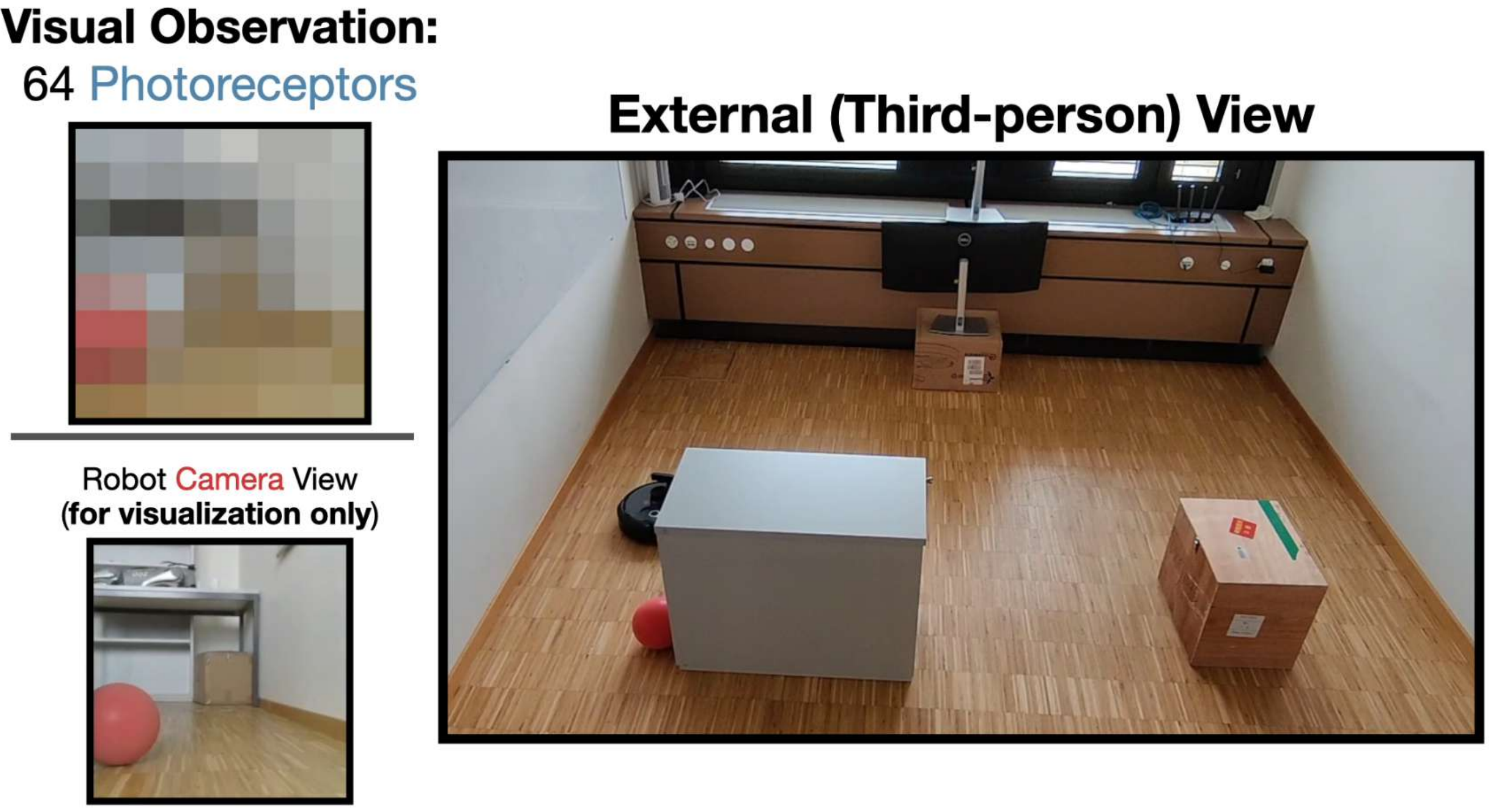}
    \caption{
    \footnotesize \textbf{Evaluation of a PR equipped agent in the real world.}
        We deployed a control policy using 64 PRs on a real robot for target navigation. It successfully navigates to the target ball in an unknown room with no real-world training.
    }
    \label{fig:real-world} 
\end{figure}

We deployed a control policy using 64 PRs (less than 1\% of the camera resolution) on a real robot as shown in \cref{fig:real-world}. It demonstrates impressive performance, successfully navigating to the target ball in an unknown room with no real-world training, relying solely on the low-resolution visual signal. The results can be seen at \href{https://visual-morphology.epfl.ch/#real-world}{https://visual-morphology.epfl.ch/\#real-world}.

\section{Discussion and Limitations}

In this work, we aim to demonstrate that even extremely simple visual sensors like photoreceptors can be effective in solving vision tasks that require an understanding of the surrounding world and the self (proprioception).
This shows that, similar to numerous examples in nature, a system with certain simplicities can exhibit intelligent and complex behaviors.
This also suggests an avenue for an interesting research direction in addition to the trends focused on training larger models on ever-increasing amounts of data and complex sensors. 

We demonstrate that design optimization of simple visual sensors is important to achieve performance similar to that of a more complex camera sensor.
We, therefore, approach this problem computationally and suggest a design optimization method that is able to improve the initial design and find well-performing designs.
Below, we discuss some limitations of our work.

\textit{\textbf{Scope of the Scenarios, Vision Tasks, and Agents.}}
We instantiated a first attempt in this area and focused on an active vision tasks primarily around locomotion and typical robotic agents.
Exploring other visual tasks would be useful to better understand the limits and applicability of simple visual sensors. Similarly, the most useful scenario for the narrative we provided may not necessarily be typical tasks and typical robots in typical environments, but rather less usual ones, e.g., a perceptual micro-robot that gets injected in the body to perform a medical task. 

\textit{\textbf{Additional Constraints and Regularized Design Optimization.}}
In this work, we primarily focused on the performance of the \prab sensor to demonstrate its effectiveness.
However, other aspects, such as the number of sensors, production costs, power consumption, or physical size constraints, matter and are important to be included in the optimization objective. For instance, the ``Square–cube law'' shows disregarding weight distribution constraints would falsely suggest that the body of an animal can grow in size to a level that is practically impossible.

\textit{\textbf{Design Space Parametrization for Complex Robot Morphology.}}
The robot morphologies considered in this work primarily consist of primitive shapes such as boxes and cylinders, which makes it relatively simple to parametrize the design space to be constrained to the robot's body.
However, many real-world robots, e.g., soft robots, might have more complex shapes.
Developing a general way to parameterize more complex robots' surfaces is a direction to make design optimization methods easily applicable.

\textit{\textbf{Local Design Optimization.}} Our design optimization method is able to improve upon initialization and find a well-performing design. 
However, starting from different initializations may be important, as it is a local optimization method that can be stuck at local optima.
Incorporating methodologies from global search methods may be useful to achieve better overall performance.

{
\small
\bibliographystyle{ieeenat_fullname}
\bibliography{references}
}


\appendix
\clearpage
\newpage


\section*{Appendix Overview}
The Appendix provides further discussions, details, and evaluations as outlined below:
\begin{itemize}
    \item In \cref{sec:dmc-state-regression}, we study whether photoreceptor sensors allow extracting information about the state of the world and whether better-performing designs lead to a more accurate world state estimation.
    \item \cref{sec:analysis} presents various analysis experiments: 1) we show that the photoreceptor agent can do target detection, 2) we show the effectiveness of the design optimization method through various ablation experiments, and 3) we perform an experimental evaluation of the importance of different design variables (such as height, pitch, etc.).
    \item \cref{sec:viz} provides additional visualizations of different designs, including random, intuitive, and computational designs and their corresponding performance.
    \item \cref{sec:dmc-add-res} provides additional results of using grids of 4x4 photoreceptors for continuous control tasks in DMC.
    \item In \cref{sec:expts-desc}, we provide a detailed description of our experimental settings. We provide details on the control policy training process and design optimization.
    \item \cref{app:intuitive-survey} provides details on the human study we developed to collect human intuitive designs for both navigation and continuous control tasks.
\end{itemize}

\section{Can photoreceptors extract information about the world state?}

\label{sec:dmc-state-regression}
In Sec. 4.2 of the main paper, we demonstrated that an agent equipped with only a few photoreceptors can perform well in solving active vision-based tasks.
One would expect such a PR agent to be able to extract useful information about the state of the world using its visual sensors.
In this section, we explore whether photoreceptors can extract information about the state of the world and self, and whether better-performing designs extract state information more accurately.
\begin{figure}
    \centering
    \includegraphics[width=\linewidth]{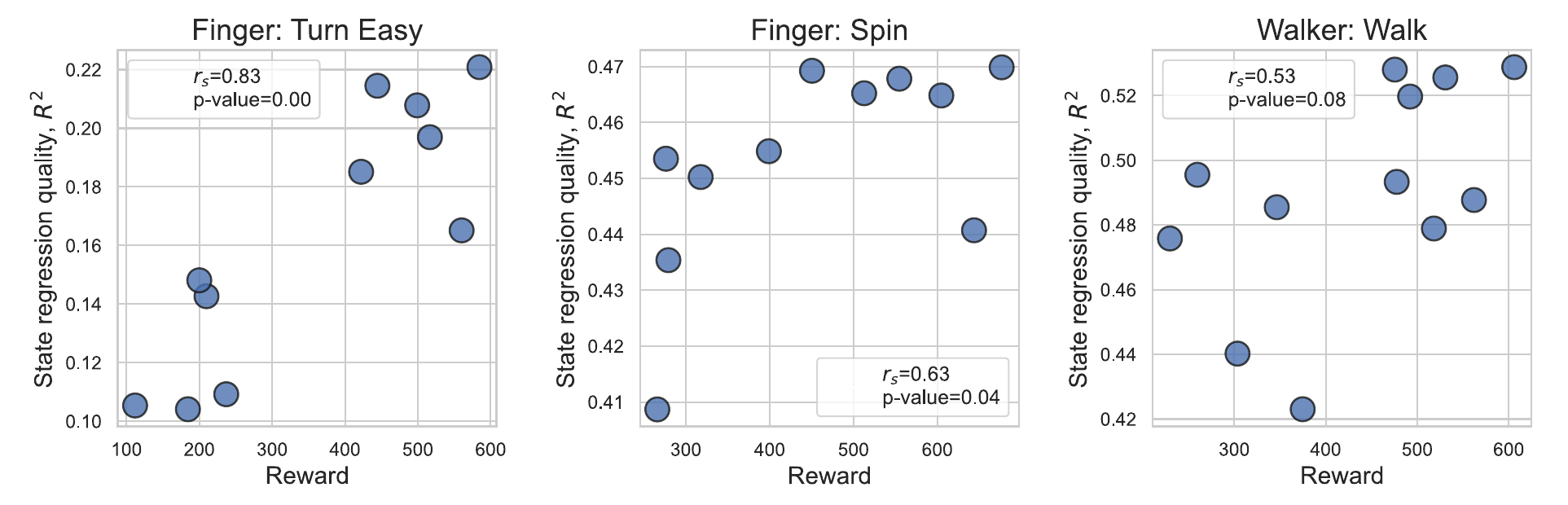}
    \caption{
    \textbf{The quality of the state estimation vs. the reward for different photoreceptor designs}.
    For each design of the photoreceptor sensors, $y$-axis shows the quality of the state regression from the visual signal provided by this design as measured by the coefficient of determination $R^2$ averaged along all available state variables on a test set (see~\cref{sec:dmc-state-regression} for more details), and $x$-axis shows the reward achieved by the agent using this design.
    We use different designs (random, computational, and intuitive) for 1x1 photoreceptors with $K\in\{2, 4\}$.
    First, we find that $R^2 > 0$ in all cases, meaning that there is useful information about the state contained in the signal of the photoreceptor sensors.
    More interestingly, we find evidence for the correlation between state regression quality and reward.
    \textit{This suggests that the quality of the state regression can be a good proxy for finding a well-performing design in terms of the reward on the active task.}
    }
    \label{fig:dmc-regress}
\end{figure}

\begin{figure}
    \centering
    \includegraphics[width=\linewidth]{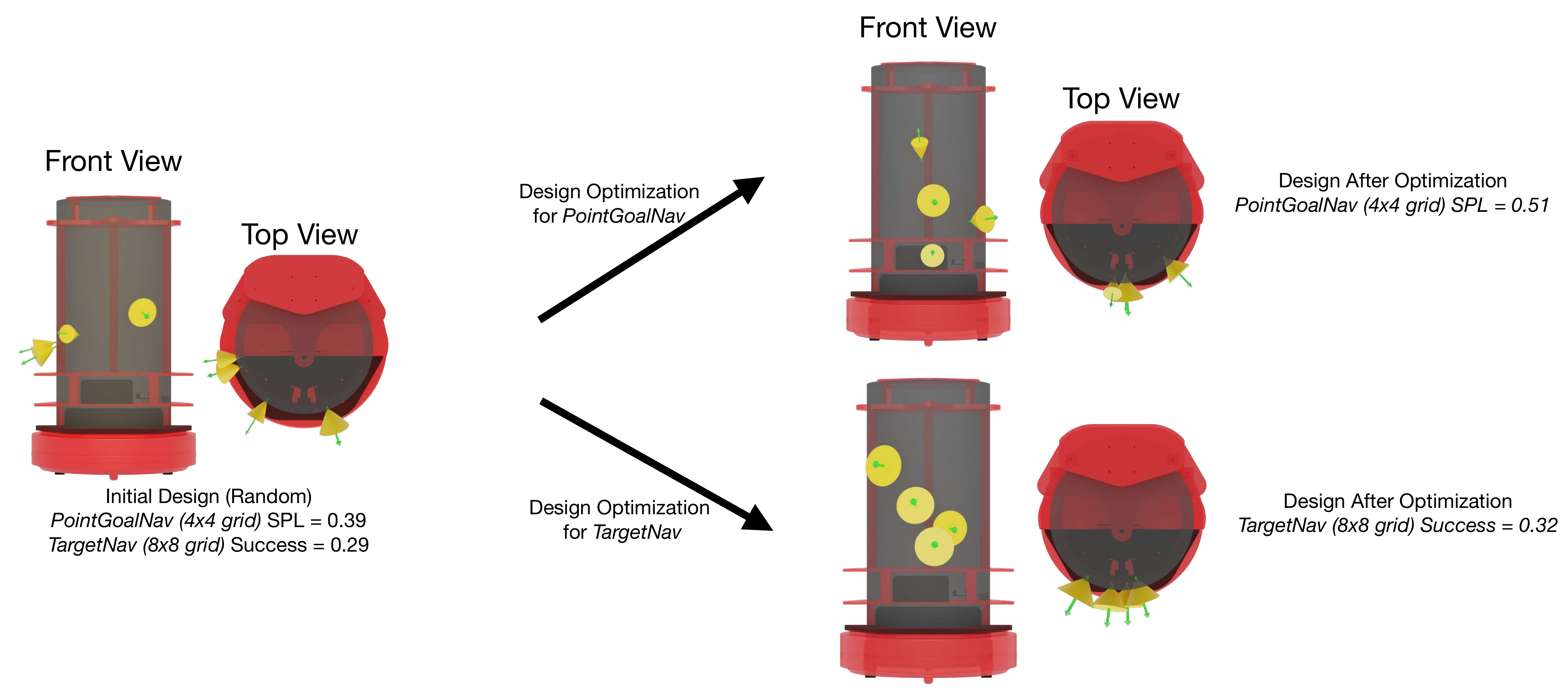}
    \caption{\textbf{Computational design obtained on \texttt{PointGoalNav} and \texttt{TargetNav} staring from the same initial random design.} The proposed design optimization method produces different designs for the navigation tasks while improving performance on the specific task. Each performance value (Success for \texttt{TargetNav} or SPL for \texttt{PointGoalNav}) is obtained by training a control policy with the design fixed as shown in the images.}
    \label{fig:comp-design-compare}
\end{figure}

We consider three tasks from the DMC Suite: \texttt{Finger: Spin}, and \texttt{Finger: Turn Easy} and \texttt{Walker: Walk}.
For each task, we collect rollouts using the best-performing policy available.
At each step, we collect the default state information provided by the DMC benchmark.
For example, for the \texttt{Walker: Walk} task, this includes the height of the body, orientations, and velocities of each body part.
These state values are the default variables used as input by the state-based control algorithms and, therefore, provide a sufficient description of the world and agent states.

In addition to the state information, we collect visual sensory data for different designs (random, computational, and intuitive), achieving different reward values.
Then, for each design, we regress the state values from the visual sensory data using the same backbone as for the policy network (we train it from scratch).
We use 80000 timestamps for training and 20000 for testing (test time stamps come from different episodes).
For each state variable, we measure the coefficient of determination $R^2$ on the test set and average it over all state dimensions, representing the overall quality of the state estimation from the photoreceptor sensors with the corresponding design.

\cref{fig:dmc-regress} shows the quality of the state estimation versus the reward for each design (we use 1x1 photoreceptors with $K\in\{2,4\}$).
First, we find that the $R^2$ is greater than zero, which means that it is possible to extract more information about the state than the overall mean value (since $R^2=0$ corresponds to a mean prediction).

We also find evidence for the correlation between the quality of the state regression and the performance of the agent with the corresponding design.
This suggests the quality of the state regression can be a good proxy for a design optimization method.
This is useful because this proxy represents a supervised learning task for which a design optimization can be easier to perform than directly optimizing the performance on the active reinforcement learning task.

\section{Analysis Experiments and Ablations}
\label{sec:analysis}

\subsection{Does the task affect the computationally obtained design?}
\cref{fig:comp-design-compare} shows that even with the same initial random design, the proposed design optimization method converges to different designs for different tasks, namely, \texttt{PointGoalNav} and \texttt{TargetNav}.

\subsection{Design optimisation method (computational design) uses available sensors well}

\begin{figure}
    \centering
    \includegraphics[width=\linewidth]{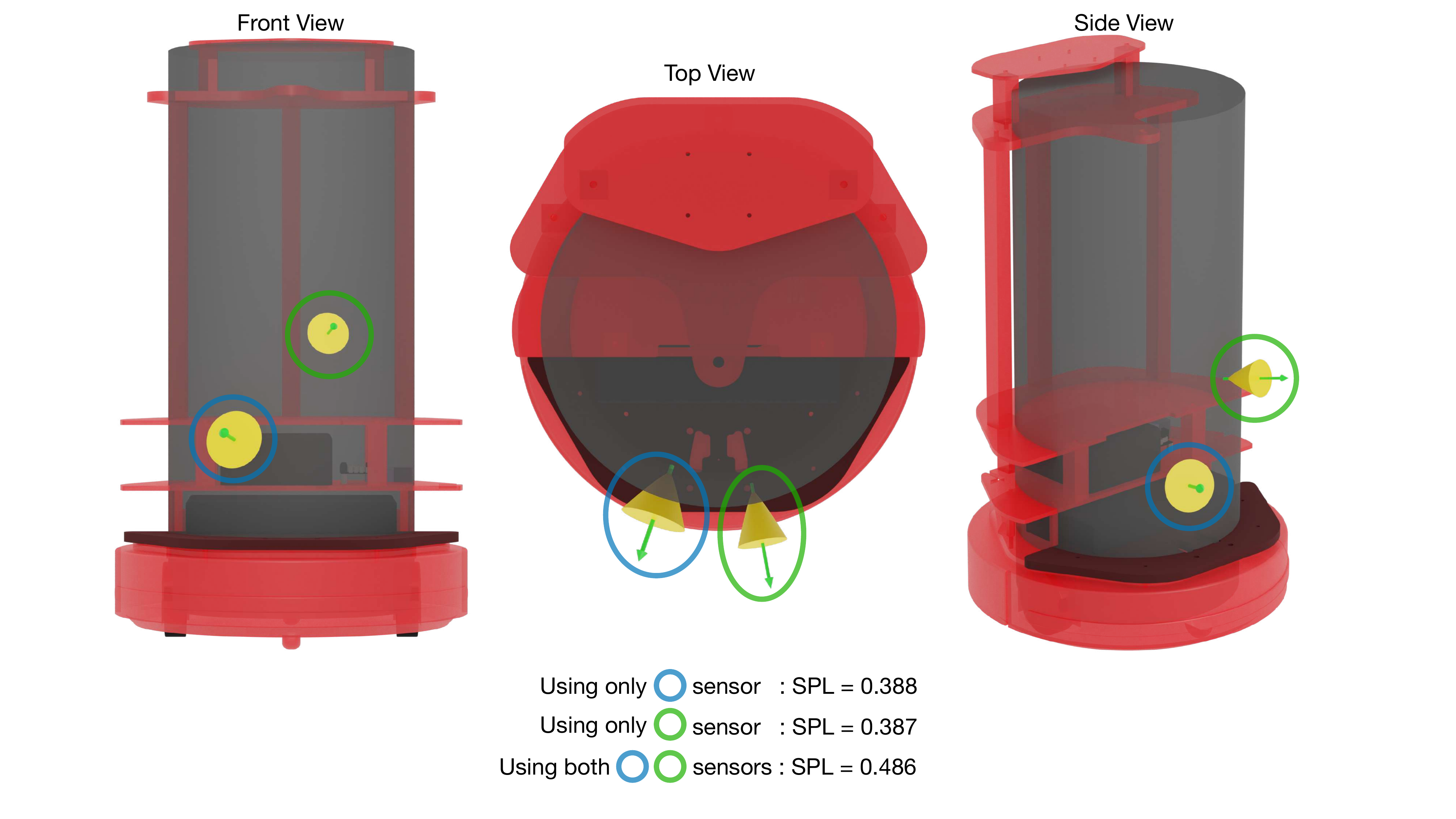}
    \caption{\textbf{Ablation of sensor grids shows that the proposed design optimization method uses available photoreceptor sensors well.} In \texttt{PointGoalNav}, designs using only one of the two sensor grids from the computational design perform significantly worse than when using both sensor grids. This shows that the design optimization method places the sensor grids so that the information from both is complementary.}
    \label{fig:dab}
\end{figure}

We run a design ablation to show that the proposed design optimization method is optimizing the placement of sensors to maximize performance. We choose the simplest setting of K=2 grids of 4$\times$4 photoreceptors in the \texttt{PointGoalNavigation} setting. From the computational design with K=2, we create two designs of K=1 by picking one of the two sensor grids in the computational design. In \cref{fig:dab}, we show a comparison between the original computational design and the ablated designs, showing that the design optimization utilizes the placement of the additional sensor grid effectively as neither of the two sensor grids alone performs well but together the performance is significantly boosted.

\subsection{Photoreceptor-based agents can do Target Detection}

\begin{figure}
\centering
\begin{subfigure}{.5\linewidth}
  \centering
  \includegraphics[width=0.9\linewidth]{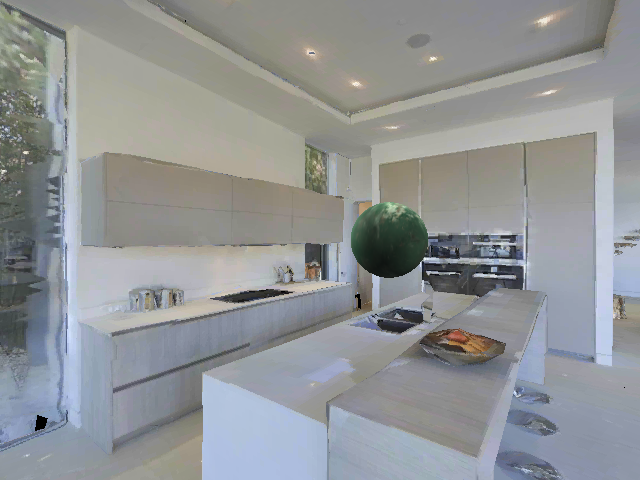}
\end{subfigure}%
\begin{subfigure}{.5\linewidth}
  \centering
  \includegraphics[width=.9\linewidth]{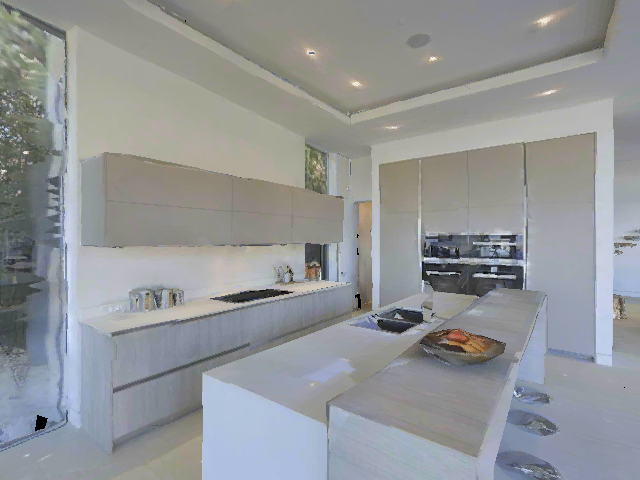}
\end{subfigure}
\caption{\textbf{Comparison between environments with the green target sphere and with a transparent sphere.} Left: An environment with the green target sphere. Right: An environment with a transparent sphere. Through the comparison, we can find that the green sphere does not influence the view of the surrounding objects. Thus, the surrounding objects do not provide a hint about where the target is. The photoreceptor agent has to identify the target ball to successfully navigate to it.}
\label{fig:compare_with_without_target_ball}
\end{figure}

\begin{figure}
    \centering
    \includegraphics[width=\linewidth]{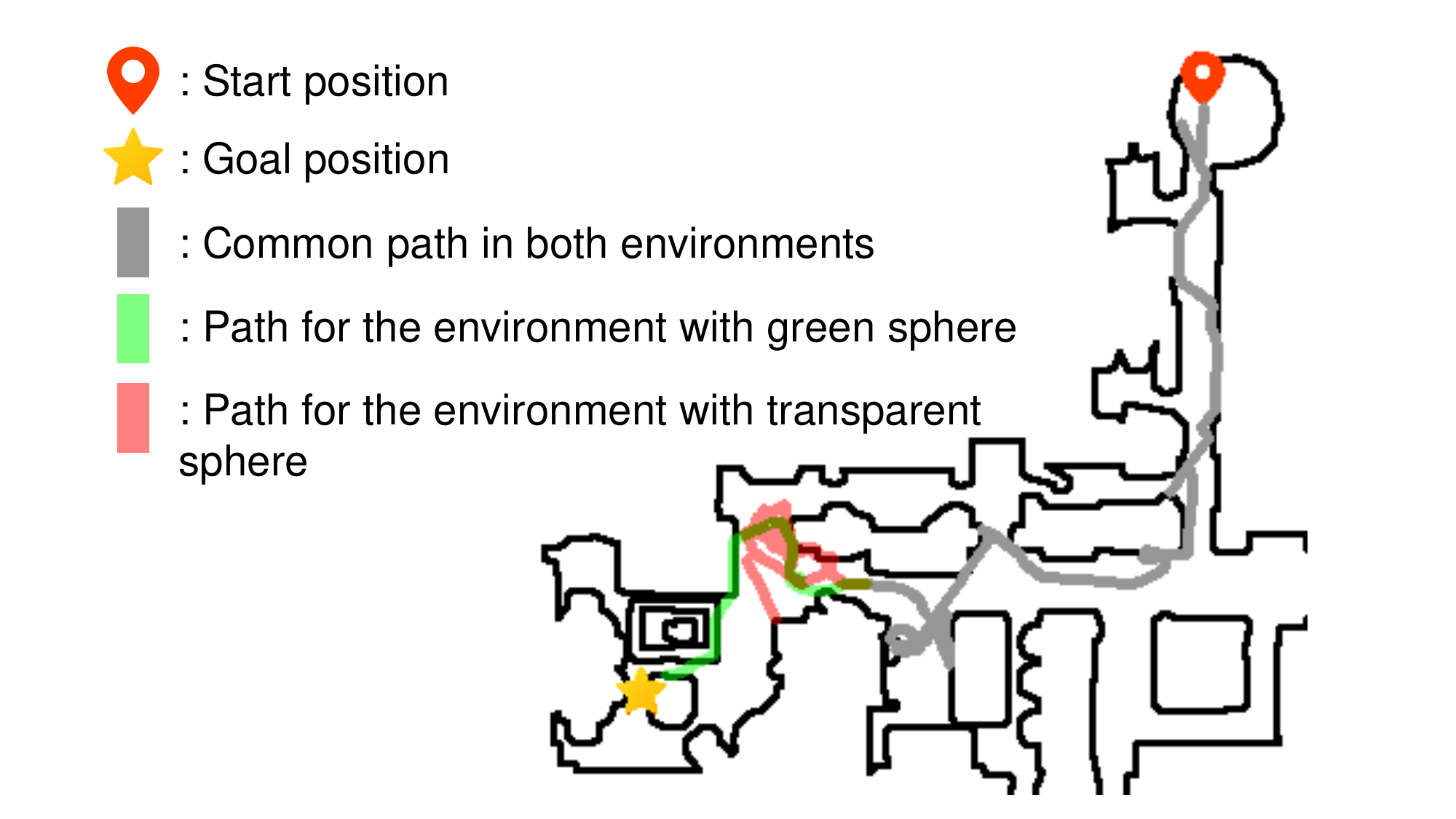}
    \caption{\textbf{Photoreceptor agent successfully navigates to the green target sphere while continues searching when the target sphere is transparent.} This shows that the PR agent is capable of target detection.}
    \label{fig:diverge_compose}
\end{figure}

In the \texttt{TargetNav} task, to confirm that photoreceptors can perform target detection, i.e., identify the green sphere and move towards it, we test the behavior and performance of the trained PR agent with a transparent sphere as the target instead of the green sphere. The comparison between an environment with the green sphere and the transparent sphere is shown in \cref{fig:compare_with_without_target_ball}. \cref{fig:diverge_compose} shows trajectory visualizations comparing the two settings: one with a green target sphere and the other with a transparent sphere in otherwise identical episodes. We see that initially, in both cases, the PR agent follows the same trajectory. In the episode with the green target, the PR agent is able to recognize it and move towards it, while in the episode with the transparent target, the agent does not see it (as expected) and continues searching for it. For a quantitative comparison, to demonstrate that the PR agent is indeed performing target detection and moving towards it to achieve success, rather than only conducting efficient exploration, we compare the agent's success rate in both target settings. \cref{tab:perf_with_without_gree_sphere} shows that the PR agent is indeed performing target detection, resulting in a much higher success rate.

\begin{table}[]
\centering
\begin{tabular}{ccc}
\toprule
             & \begin{tabular}[c]{@{}c@{}}With \\ green sphere\end{tabular} & \begin{tabular}[c]{@{}c@{}}With \\ transparent sphere\end{tabular} \\ \hline
Success Rate & 0.314                                                        & 0.132        \\                                                     \bottomrule
\end{tabular}
\vspace{5pt}
\caption{\textbf{Photoreceptor agents achieve a higher success rate when using the green sphere as the target compared to using the transparent sphere as the target.}
The gap between PR agent performance in the environment with the green sphere as the target and that using the transparent target proves that the PR agent can identify the target ball and navigate to it.
}
\label{tab:perf_with_without_gree_sphere}
\end{table}

\subsection{Comparing importance of the different design space variables}
\begin{figure}
    \centering
    \includegraphics[width=0.8\linewidth]{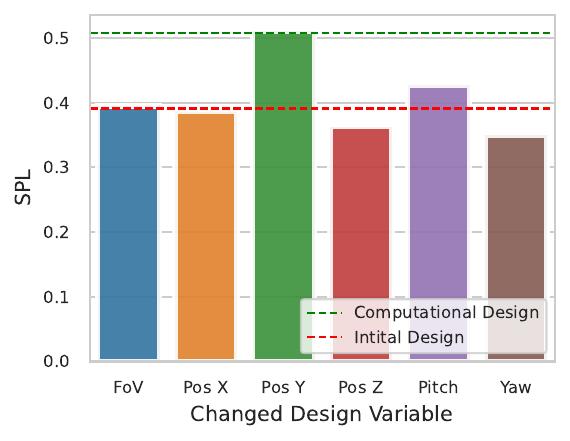}
    \caption{\textbf{Comparison of usefulness of different design axes.} We evaluate the performance of each design created by starting from the computational design, and for all PR grids, setting all but one design variables to their initial values (before optimization).}
    \label{fig:change}
\end{figure}

We run additional experiments in order to compare the importance of the different design variables in the design space defined in the main manuscript, i.e.,  $[x_i, y_i, z_i, \yaw_i, \pitch_i, \fov_i]$. For measuring the importance of a specific design variable, $x_i$ for example, starting from the computational design, for all PR grids, we set all design variables except $x_i$ to their initial values (before optimization). This comparison between different design axes is shown in \cref{fig:change} for K=2 grids of 8$\times$8 PRs in \texttt{PointGoalNavigation}, which shows that the height $y_i$ and the pitch $\pitch_i$ design variables are the most important. We also show visualizations for each of these design change in \cref{fig:change-viz}.

\subsection{Distorting the computational design to analyse the success of design optimisation}
\begin{figure}
    \centering
    \includegraphics[width=0.8\linewidth]{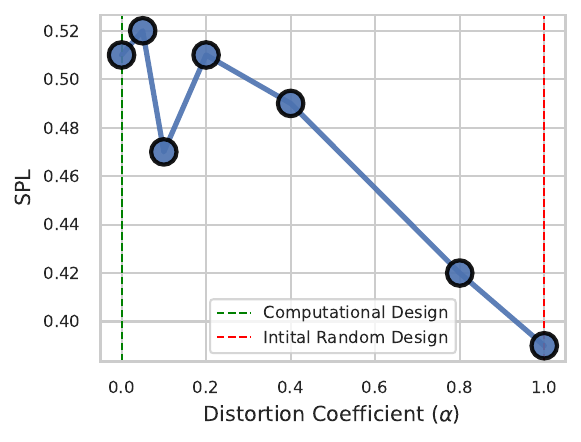}
    \caption{\textbf{Designs farther away from the computational design perform worse.} We conduct a design distortion experiment in the \texttt{PointGoalNav} task where we compute interpolated designs between the computational and initial designs using the formula : $\theta_{interpolated} = (1-\alpha) \times \theta_{Computational} + \alpha \times \theta_{Initial}$, choosing an exponentially increasing $\alpha$. The plot shows that designs farther away from the computational designs perform much worse.}
    \label{fig:distort}
\end{figure}

\begin{figure*}
    \centering
    \includegraphics[width=0.7\textwidth]{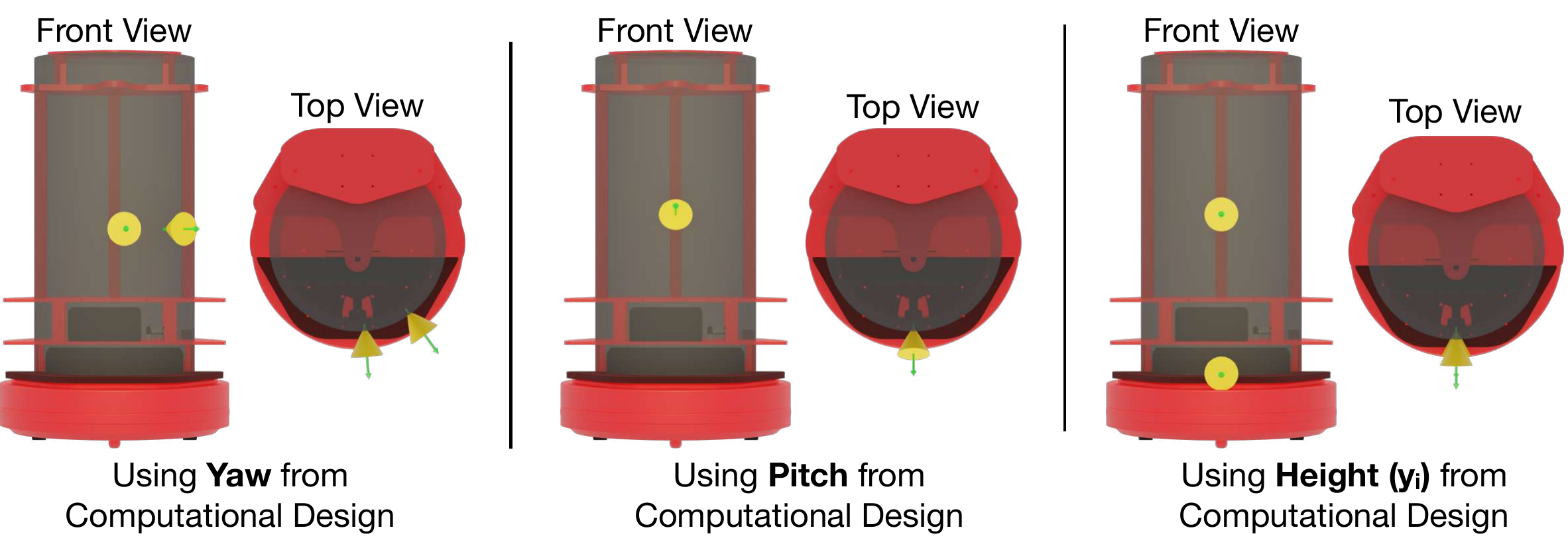}
    \caption{\textbf{Visualization of designs comparing importance of different design axes.} Each design is created by starting from the computational design, and for all PR grids, we set all design variables except one to their initial values (before optimization).}
    \label{fig:change-viz}
\end{figure*}

For probing the design optimization landscape, we create designs by interpolating between the computational and initial designs using the following formula: \\
\begin{equation}
    \theta_{interpolated} = (1-\alpha) \times \theta_{Computational} + \alpha \times \theta_{Initial}
\end{equation}
We choose an exponentially increasing distance from the computational design for the interpolation, i.e., $\alpha \in \{0.05, 0.1, 0.2, 0.4, 0.8\}$ and train control policies for the obtained designs. The performance obtained for each such design is shown in \cref{fig:distort}.

\section{Design Visualizations}
\label{sec:viz}
In this section, we provide additional visualizations of the designs we obtained through computational optimization, intuitive survey or random sampling and their respective performance on the corresponding task.

\subsection{Computational vs Random Design Visualisations}

\begin{figure}
    \centering
    \includegraphics[width=\linewidth]{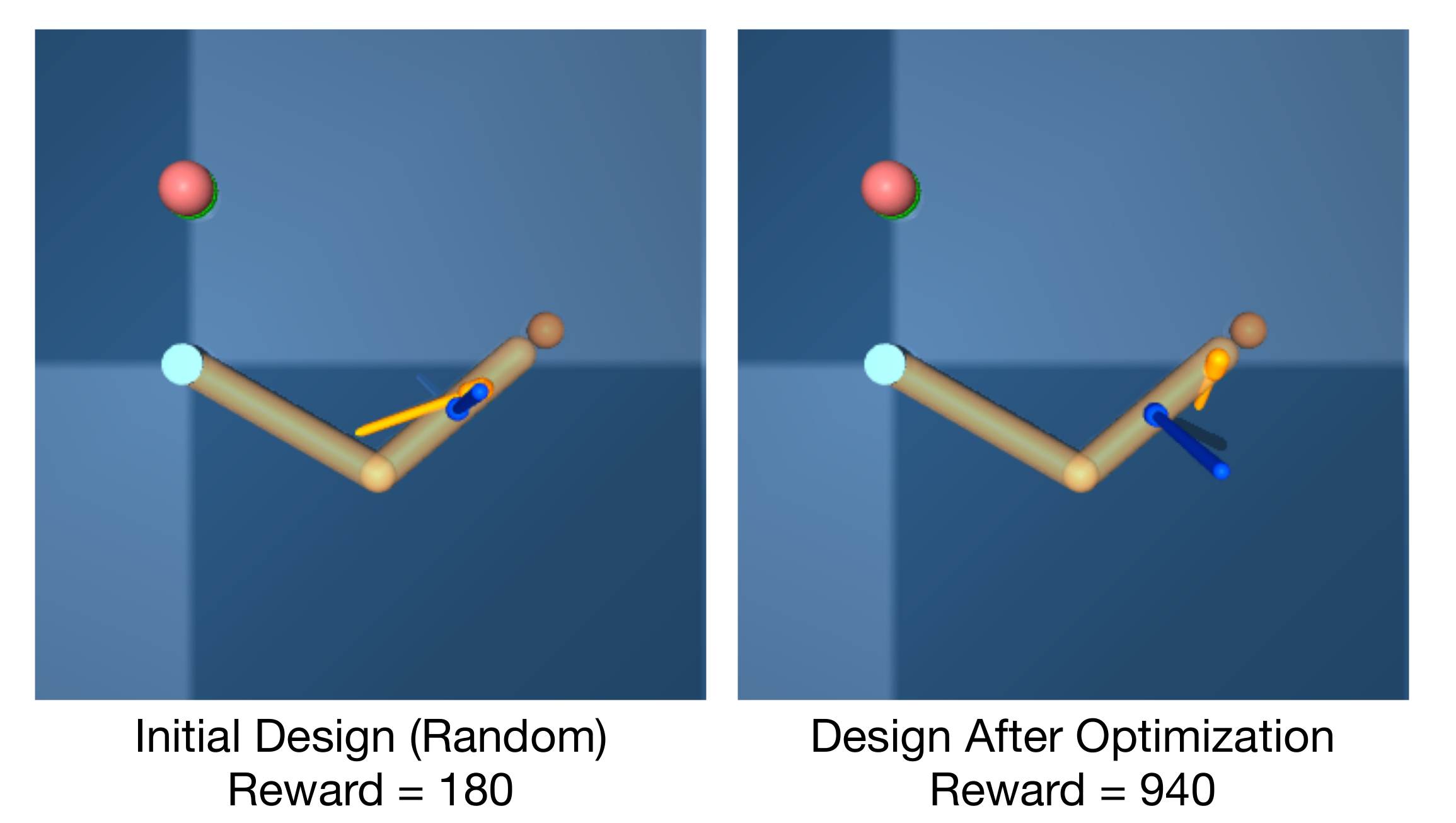}
    \caption{\textbf{Visualisation and performance comparison on \texttt{Reacher-Hard} task} of initial random and computationally (through optimisation) obtained designs.}
    \label{fig:dopt-reacher}
\end{figure}

\begin{figure}
    \centering
    \includegraphics[width=\linewidth]{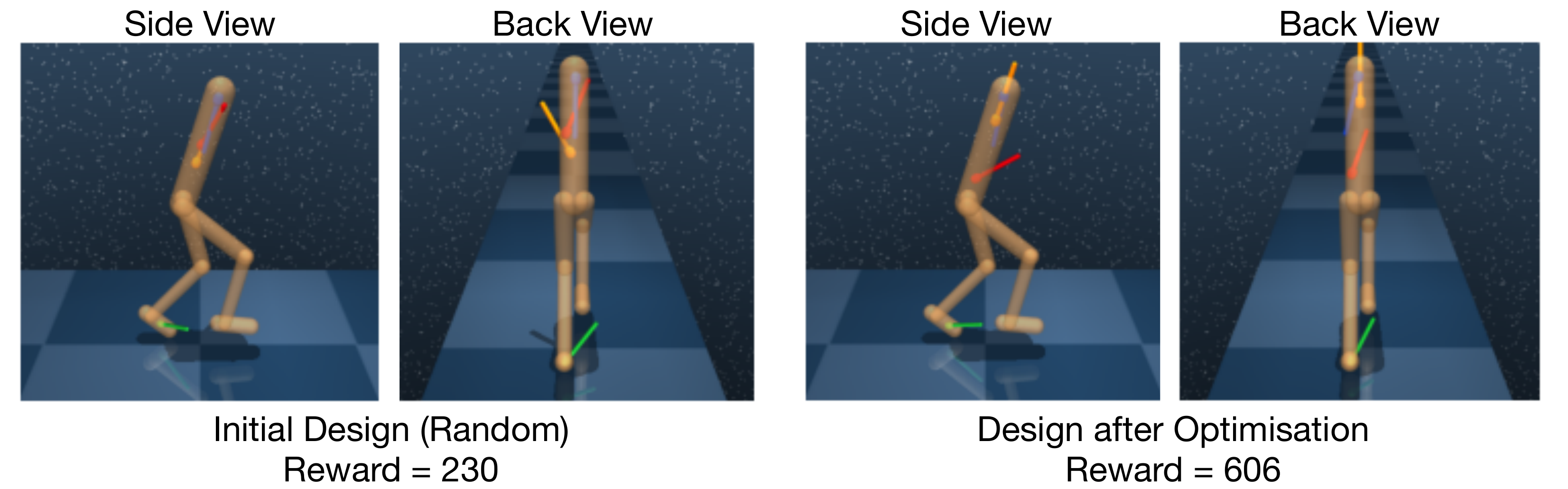}
    \caption{\textbf{Visualisation and performance comparison on \texttt{Walker-Walk} task} of initial random and computationally (through optimisation) obtained designs.}
    \label{fig:dopt-walker}
\end{figure}

\begin{figure}
    \centering
    \includegraphics[width=\linewidth]{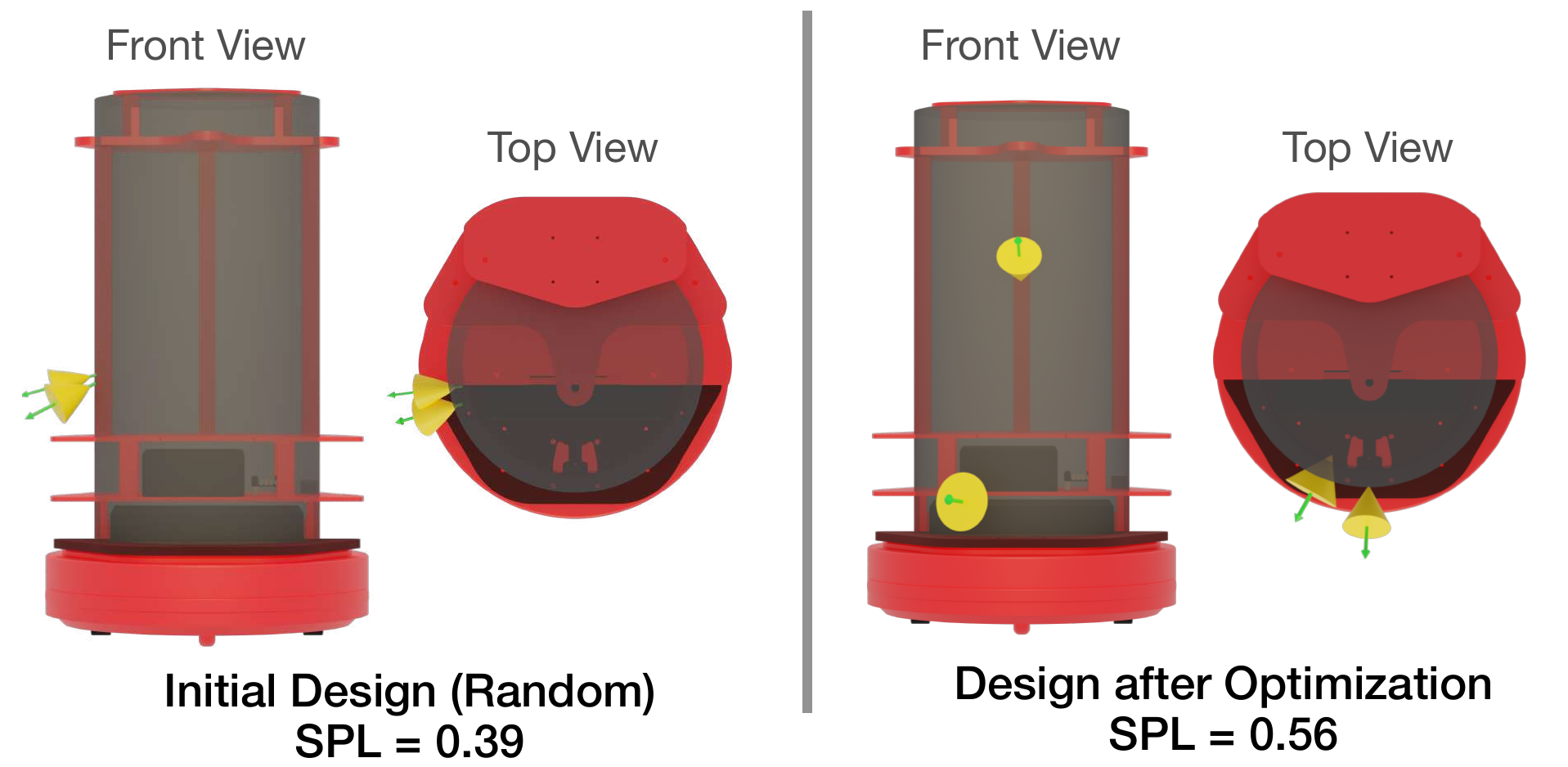}
    \caption{\textbf{Visualisation and performance comparison on \texttt{PointGoalNav} task} of initial random and computationally (through optimisation) obtained designs.}
    \label{fig:dopt-ptnavk4-4x4}
\end{figure}

In \cref{fig:dopt-reacher} and \cref{fig:dopt-walker}, we show the initial random design and the corresponding computational design obtained using the proposed design optimization method for DeepMindControl and \cref{fig:dopt-ptnavk4-4x4} shows the same for the \texttt{PointGoalNav} task. The figures also show the improved reward corresponding to the computational design.

\subsection{Intuitive Design Visualisations}
\begin{figure*}
    \centering
    \includegraphics[width=0.8\textwidth]{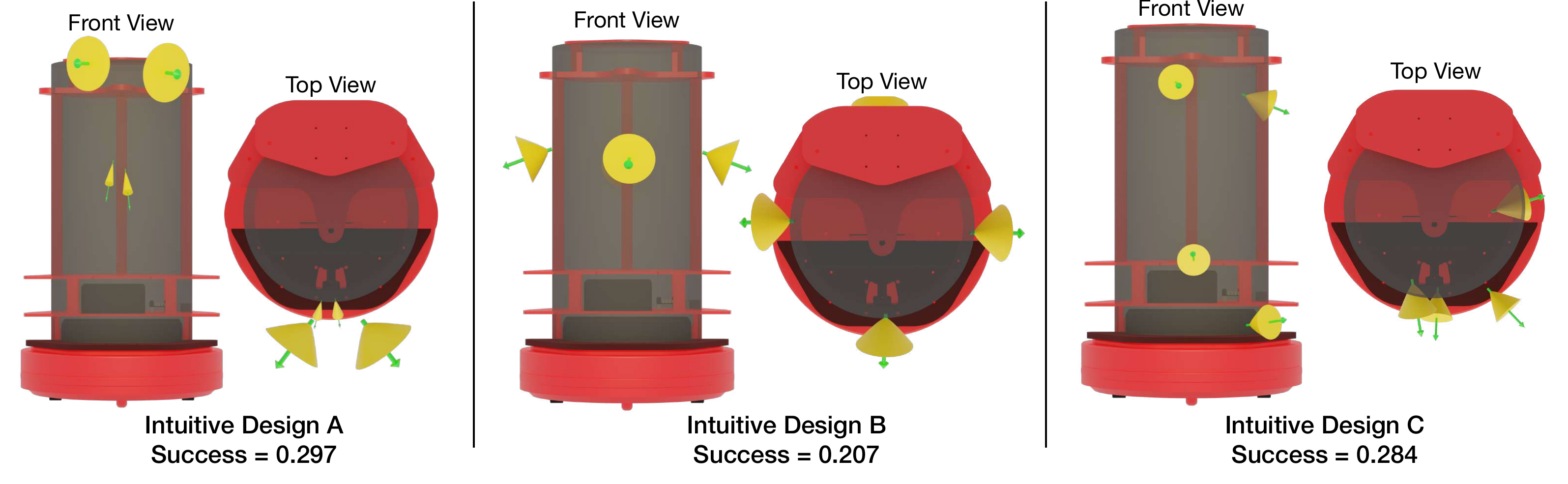}
    \caption{\textbf{Visualisation and performance comparison on \texttt{TargetNav} task} of different intuitive designs collected using the human study survey describe in \cref{app:intuitive-survey}.}
    \label{fig:tgtnav-int}
\end{figure*}

\cref{fig:tgtnav-int}, shows visualizations for some of the intuitive designs collected using the survey described in \cref{app:intuitive-survey} and their corresponding performance on the \texttt{TargetNav} task. This shows that the variance in the performance of intuitive designs is high as well.

\section{Additional Results for Continuous Control Tasks using the Grids of 4x4 Photoreceptors}
\label{sec:dmc-add-res}
\begin{figure*}[t]
    \centering
    \includegraphics[width=.9\textwidth]{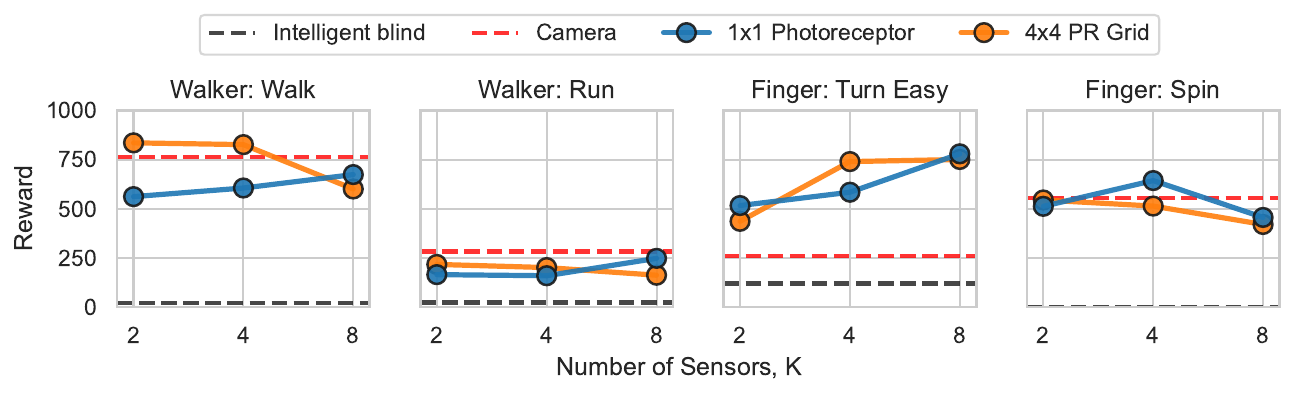}
    \caption{
    \textbf{Additional results for DMC using grids of 4x4 photoreceptors.}
    For the four most difficult tasks (where neither agent achieves near-optimal performance, i.e., the reward of 1000), we explored using more sensors by arranging PRs in the 4x4 grids similar to the navigation tasks (see Sec.~3.1 of the main papers for more details).
    We find that it considerably improves the performance for the \texttt{Walker: Walk} task and either does not improve or even deteriorates the performance on other tasks.
    }
    \label{fig:dmc-4x4}
\end{figure*}

In addition to the results in Fig.~4 for using 1x1 photoreceptors, we explore whether using a grid of 4x4 photoreceptors further improves the performance of the PR agents.
\cref{fig:dmc-4x4} presents the results on the four most difficult tasks (i.e., where neither agent achieved high performance close to the optimal reward of 1000).


\section{Experimental Details}
\label{sec:expts-desc}
\subsection{PointGoal Navigation Setting}
\label{sec:pointnav_setting}

In \texttt{PointGoalNav}, the agent is randomly initialized in an environment and asked to navigate to a target point given relative to the start state.
The episode ends if the agent calls the stop action. It succeeds in the episode if it stops within a 0.2-meter radius of the target point.

\textbf{Observation Space.}
The agent has access to an idealized \texttt{GPS+Compass} input that provides its current position and rotation \textit{relative} to the starting state. 
It also receives the relative position of the target point.
In addition, the agent observes egocentric RGB views through its photoreceptors.

\textbf{Action Space.}
The agent can execute 4 actions: \texttt{move$\_$forward} (0.25m), \texttt{turn$\_$left} (30$^\circ$), \texttt{turn$\_$right} (30$^\circ$), and \texttt{stop}.

\textbf{Reward.}
At every timestep $t$, the agent at state $s_t$ has the geodesic (shortest path) distance $d_t$ to the target. 
It applies an action $a_t$ and transits to the next state $s_{t+1}$ whose geodesic distance to the goal is $d_{t+1}$. 
It receives a reward $r_t$ in the form of
\begin{equation*}
    r_t=
    \begin{cases}
        2.5 * \text{Success} & \text{if } a_t \text{ is \texttt{stop}} \\
        d_t-d_{t+1}-c^\text{slack} & \text{otherwise}
    \end{cases}
\end{equation*}
where $d_t-d_{t+1}$ is a dense reward for progressing towards the target position and $c^\text{slack}=0.003$ is the slack reward encouraging shorter episodes.

\textbf{Dataset.} In \texttt{PointGoalNav}, we use the Matterport3D dataset \cite{chang_matterport3d_2017} for training and testing.
The training data (\texttt{train} split) contains 61 scenes with around 80k episodes per scene.
The testing data (\texttt{test} split) contains 18 scenes \textit{unseen} during training with 56 episodes per scene (in total 1008 episodes).

\textbf{Training Process.}
We train our navigation agents with Proximal Policy Optimization (PPO)~\cite{schulman_proximal_2017} which optimizes the objective 
\begin{equation*}
    L(\omega)=\mathbb{E}[\min(r_t(\omega)\hat{A}_t, \text{clip}(r_t(\omega), 1-\epsilon, 1+\epsilon)\hat{A}_t)]
\end{equation*}
where $\omega$ parameterizes the control policy, $r_t(\omega)$ is the probability ratio between the current policy and the rollout policy, and $\hat{A}_t$ is the estimate of the advantage function through Generalized Advantage Estimation (GAE)~\cite{schulman_high-dimensional_2018}. 
In practice, we adopt the training mechanism in Decentralized Distributed PPO (DD-PPO)~\cite{wijmans_dd-ppo_2020} to accelerate the training.
We train the agent on 4 A100 or V100 GPUs while each GPU has 30 parallel environments collecting 64 steps of experience (simulation steps) per environment. 
We call collecting the rollouts above once as a \textit{rollout collection step} ($\approx$ 7.6k simulation steps).
With the collected rollouts, we perform 4 epochs of PPO update with 1 mini-batch per epoch.
We use the Adam~\cite{kingma_adam_2017} optimizer with an initial learning rate of $2.5\times 10^{-4}$. 
We set the clipping parameter $\epsilon$ to 0.2, the discount factor $\gamma$ to 0.99, the GAE hyperparameter $\lambda$ to 0.95. We train the agent for around 230 million (M) simulation steps.

\textbf{Evaluation Process.}
We evaluate our agent based on the Success weighted by Path Length (SPL) metric~\cite{anderson_evaluation_2018}. 
One episode is counted as success only when the agent takes the \texttt{stop} action within 0.2 meters of the goal position within 500 steps.
The SPL reported is the average across all episodes in the \texttt{test} split.

\subsection{Target Navigation Setting}

In \texttt{TargetNav}, the agent is spawned at a random ground location in an environment and needs to navigate to a target green sphere placed randomly in the scene.
The radius of the target sphere is 0.5 m.
The agent succeeds in the episode if it enters a circle of 0.8-meter radius around the target sphere center.

\textbf{Observation Space.}
The agent receives its current position and rotation \textit{relative} to the starting point and orientation from an idealized \texttt{GPS+Compass} sensor. 
Besides, the agent also observes egocentric RGB views through its photoreceptors. 
Compared to \texttt{PointGoalNav}, the agent does not receive the target position information.
Thus, \texttt{TargetNav} focuses more on evaluating the agent's ability to explore the environment.

\textbf{Action Space.}
The agent has access to 3 actions: \texttt{move$\_$forward} (0.25m), \texttt{turn$\_$left} (30$^\circ$), and \texttt{turn$\_$right} (30$^\circ$).

\textbf{Dataset.} In \texttt{TargetNav}, we construct our training and testing data from the \texttt{PointGoalNav} dataset in Matterport3D scenes \cite{chang_matterport3d_2017}. 
We randomly sample 10 scenes from the \texttt{train} split of the \texttt{PointGoalNav} setting.
We use the same 18 test scenes as in the \texttt{test} split of the \texttt{PointGoalNav} dataset.
We use all the episodes of the \texttt{PointGoalNav} dataset for the scenes chosen above.
For each episode, we add the green sphere at a height of 1.5 meters above the ground at the goal position.

\textbf{Reward and Training Process.} 
\texttt{TargetNav} uses the same reward design and training process as in the \texttt{PointGoalNav} setting described in \cref{sec:pointnav_setting}.

\textbf{Evaluation Process.} 
In \texttt{TargetNav}, we evaluate our agent based on Success. 
One episode is successful only when the agent enters the circle of 0.8-meter radius around the target sphere center within 1500 steps.
We do not require the agent to call the \texttt{stop} action because we want to focus more on evaluating the PR agent's exploration ability using its onboard PRs.
The Success reported is the average across all episodes in the testing data.

\subsection{Continuous Control in DeepMind Control Suite}
\label{app:dmc-setting}
We use six continuous control tasks from the DeepMind Control Suite~\cite{tassa_dm_control_2020} from the following three domains:
\begin{itemize}
    \item \texttt{Reacher.} We use the difficulty level \texttt{Hard}, which requires controlling the two-legged actuator to reach the target ball with the tip of the actuator.
    \item \texttt{Walker} requires controlling a planar walker. The \texttt{Stand} task requires keeping the torso upright at some minimal height. \texttt{Walk} and \texttt{Run} tasks require, in addition, to have a specific forward velocity.
    \item \texttt{Finger} requires controlling a simple manipulator to manipulate an unactuated spinner. In the \texttt{Spin} task, the manipulator needs to spin the spinner with a specific angular velocity. In the \texttt{Turn Easy} task, one tip the of spinner needs to align with the target position specified visually.
\end{itemize}
We refer the reader to the original work \cite{tassa_dm_control_2020} for a more detailed description of action space and reward definition. 
For the \texttt{Reacher: Hard} task, we added another green target object inside the original one.
We do this because the MuJoCo renderer does not render the target ball (or any object) when the camera is inside it, and it would be impossible to realize that the camera is inside the target ball.
Therefore, we add a smaller object inside the target, which gets rendered even when the camera is inside the target and provides a visual cue for success. For all tasks, we use a common practice and repeat the same action twice (and four times for the \texttt{Finger: Turn Easy}).

\textbf{Observation Space.}
The agent only receives egocentric views from its onboard photoreceptors.
Since visual observation does not provide full information about the state (e.g., velocities), we use the standard practice of stacking three consecutive frames and using them as input to the control policy.

\textbf{Training Process.}
To maintain consistency with the navigation experiments, we use the PPO~\cite{schulman_proximal_2017} learning algorithm with the following hyperparameters.
We use $\gamma=0.99$ for reward discounting, GAE $\lambda=0.95$, and $\epsilon=0.2$ for the PPO clipping loss.
We train the control policies using half of a V100 or A100 GPU.
During training, we have 10 parallel environments while each environment collects 10000 steps of experience per rollout.
Here each rollout collection step is equivalent to 0.1M simulation steps.
We split the collected rollouts into mini-batches of 1000 and perform 4 epochs of PPO updates.
We show the specific number of simulation steps for each task in~\cref{tab:dmc-steps}.
We use Adam~\cite{kingma_adam_2017} optimization method with a learning rate $0.0001$.

\begin{table}[t]
\centering
\caption{
\textbf{The total number of steps (in million) of experiences collected for the DMC setting.} The first row (design-specific training) shows the number of steps (in million) for a fixed design. 
The second row (design optimization) shows the number of steps done (\underline{in million}) during the design optimization.
}
\label{tab:dmc-steps}
\adjustbox{max width=\linewidth}{
\begin{tabular}{@{}cccc@{}}
\toprule
\begin{tabular}[c]{@{}c@{}}Task\end{tabular} &
\begin{tabular}[c]{@{}c@{}}Design-specific\\ training\end{tabular} &
\begin{tabular}[c]{@{}c@{}}Design\\ optimization\end{tabular} \\ \midrule

\begin{tabular}[c]{@{}c@{}}Reacher: Hard\end{tabular} &
100 & 150 \\

\begin{tabular}[c]{@{}c@{}}Walker: Stand\end{tabular} &
200 & 200 \\

\begin{tabular}[c]{@{}c@{}}Walker: Walk\end{tabular} &
200 & 600 \\

\begin{tabular}[c]{@{}c@{}}Walker: Run\end{tabular} &
200 & 600 \\

\begin{tabular}[c]{@{}c@{}}Finger: Spin\end{tabular} &
200 & 800 \\

\begin{tabular}[c]{@{}c@{}}Finger: Turn Easy\end{tabular} &
300 & 800 \\ \bottomrule

\end{tabular}
}
\end{table}

\subsection{Design Optimization}
\label{app:dopt}

\textbf{Navigation Tasks.} In \texttt{PointGoalNav} and \texttt{TargetNav}, we use a Gaussian distribution as the \textit{design policy} $\pi_\phi(\theta) = \mathcal{N}(\theta \,|\,\mu, \mathrm{diag}(\sigma))$, where $\phi=(\mu, \sigma), \, \mu,\sigma \in \R^{K \times 7}$ is the mean and standard deviation, $\theta$ is the design parameter, and $K$ is the number of PRs.
We initialize the mean to be a zero vector $\mu=\bold{0}^{K\times7}$ and set the initial standard deviation to be 0.2, i.e., $\sigma=0.2\times \bold{1}^{K\times7}$. We separate the design optimization into two stages: \textit{Frozen Stage} and \textit{Update Stage}.

\textbf{Frozen Stage:} In this phase, the design policy is ``frozen'', and we only train the control policy to act as a \textit{local generalist}. At the beginning of each episode, the design parameter $\theta$ is sampled from the frozen design policy, $\theta \sim \pi_{\phi}(\cdot)$, thereby altering the robot design. As outlined in Section 5.2 of the main paper, the local generalist policy is optimized to manage control within a specific range of design parameters centered around the mean $\mu$. The scope of this range is determined by the standard deviation $\sigma$; a larger $\sigma$ allows the policy to handle a wider variety of design parameters, while a smaller $\sigma$ limits it to a narrower range. During this stage, the control policy undergoes training for 20k rollout collection steps (153M simulation steps).

{
\label{subsec:supp-dopt-update}
\textbf{Update Stage:} During this phase, both the design policy and the control policy undergo training simultaneously. We initiate updates to the design policy every 100 rollout collection steps (6M simulation steps) following each update of the control policy every 400 rollout collection steps (3.1M simulation steps).

When updating the design policy, we maintain the control policy in a frozen state. The objective is to align the design policy with the distribution of returns across the design parameter space. Instead of using returns as the primary objective, which tends to favor longer episodes due to accumulated rewards, we adopt SoftSPL (Soft Success Weighted by Path Length) \cite{datta2021integrating} as the objective function. SoftSPL balances episode efficiency and success more effectively by considering the minimum distance achieved to the target, thus providing a denser and smoother reward landscape for optimizing the design policy.

Concurrently, when updating the control policy, we freeze the design policy. This approach ensures that the control policy adapts to manage the agent within the local parameters defined by the updated design policy. Each rollout consists of 64 steps to facilitate more frequent updates of the policies.

This dual updating strategy allows for comprehensive refinement of both the design and control policies, ensuring robust performance across various task scenarios.
}

\textbf{DeepMind Control Suite.}
We use the same Gaussian design policy as in navigation setting. The total number of simulation steps dedicated to design optimization for each task is detailed in \cref{tab:dmc-steps}. During the Update Stage, we utilize return as the objective function for both the control and design policies. To adapt the control policy to the changing design policy, we update the design policy every single rollout collection step (0.1M simulation steps), following each training session of the control policy every 8 rollout collection steps (0.8 million simulation steps).

\begin{figure}
    \centering
    \includegraphics[width=0.95\linewidth]{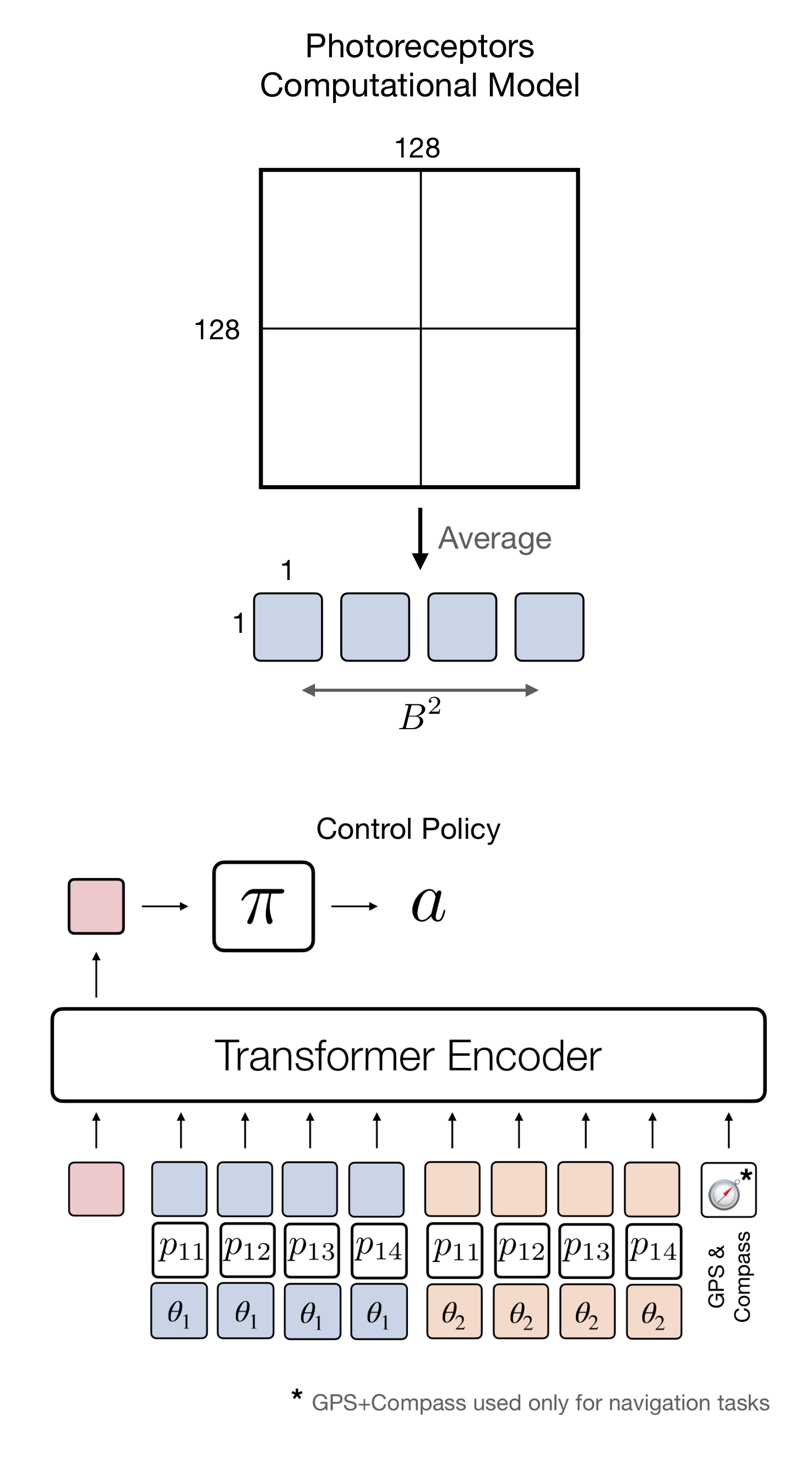}
    \caption{\textbf{Detailed figure of the transformer architecture.} Each photoreceptor triple is concatenated with the design parameters of its grid $\theta_i$ and a position embedding $p_{ij}$ according to its position on the grid. The sequence of photoreceptors along with a readout token and the GPS and compass sensor input for navigation tasks forms the input to the transformer encoder.}
    \label{fig:arch}
\end{figure}

\subsection{Network Architecture}
\cref{fig:arch} provides a detailed architecture of the transformer encoder used in both settings, i.e. Navigation and DMC. Each photoreceptor token consists of the RGB triple, the position embedding based on the position of the photoreceptor on the grid and the design parameters of the grid. This forms the encoder input to the control policy $\pi$.
In the \texttt{PointGoalNav} and \texttt{TargetNav} navigation tasks, the policy is a 2-layer LSTM, while in DeepMindControl (DMC), we stack the last 3 frames' encodings as input to the policy.

\section{Human Study for the Intuitive Designs}
\label{app:intuitive-survey}
\subsection{Visual Navigation Tasks}
To compare the performance of our design optimization algorithm with that of human engineers, we designed and conducted a survey. The survey asked participants to optimize the position, orientation, and field of view (FoV) of visual sensors on a robot, aiming to enable it to complete the \texttt{TargetNav} task as quickly as possible.

\begin{figure}
    \centering
    \begin{subfigure}[b]{0.45\textwidth}
        \centering
        \includegraphics[width=\textwidth]{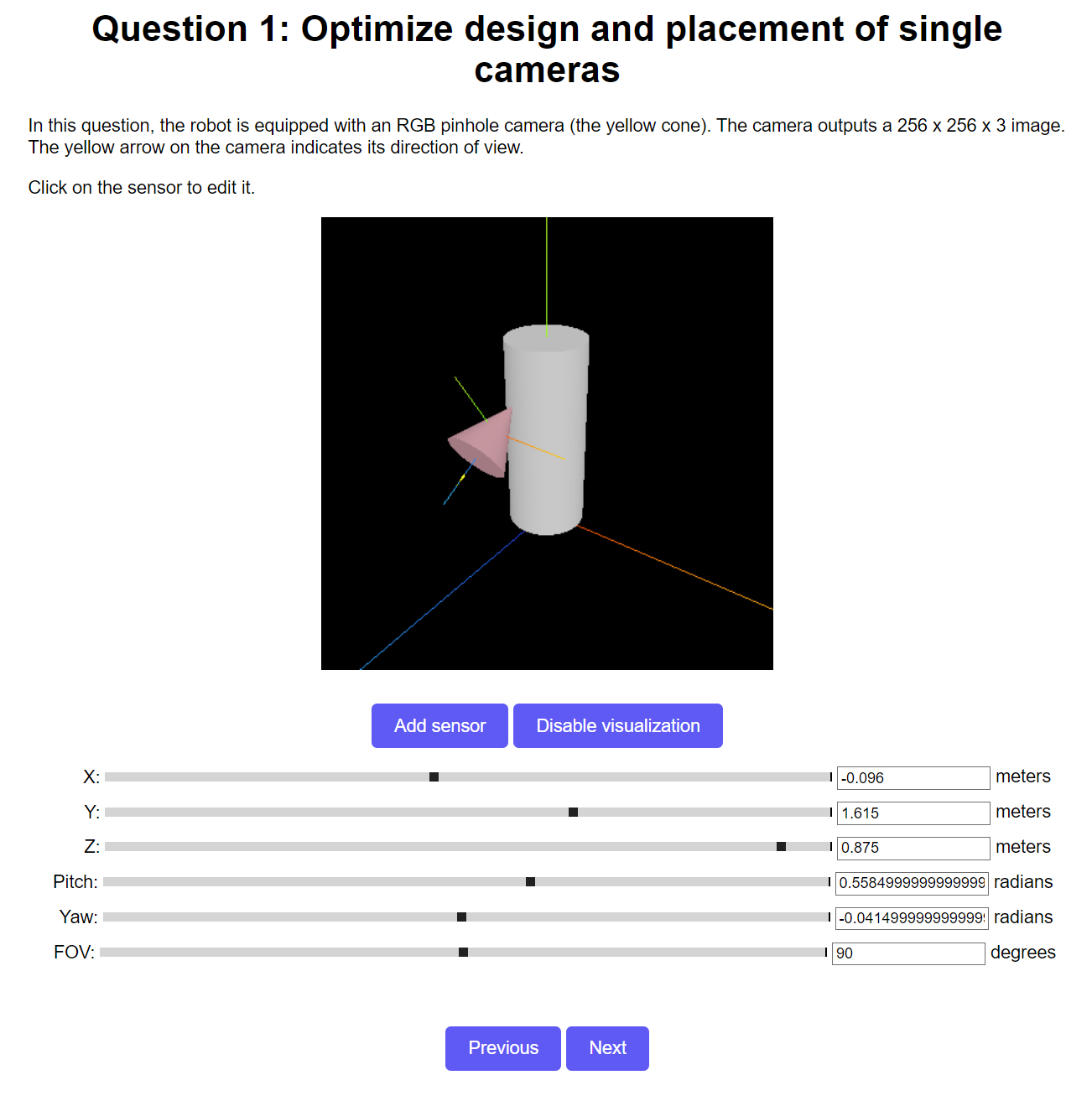}
        \caption{Level 1, question 1 (single camera design optimization).}
        \label{fig:survey-lvl1}
    \end{subfigure}
    \hfill
    \begin{subfigure}[b]{0.45\textwidth}
        \centering
        \includegraphics[width=\textwidth]{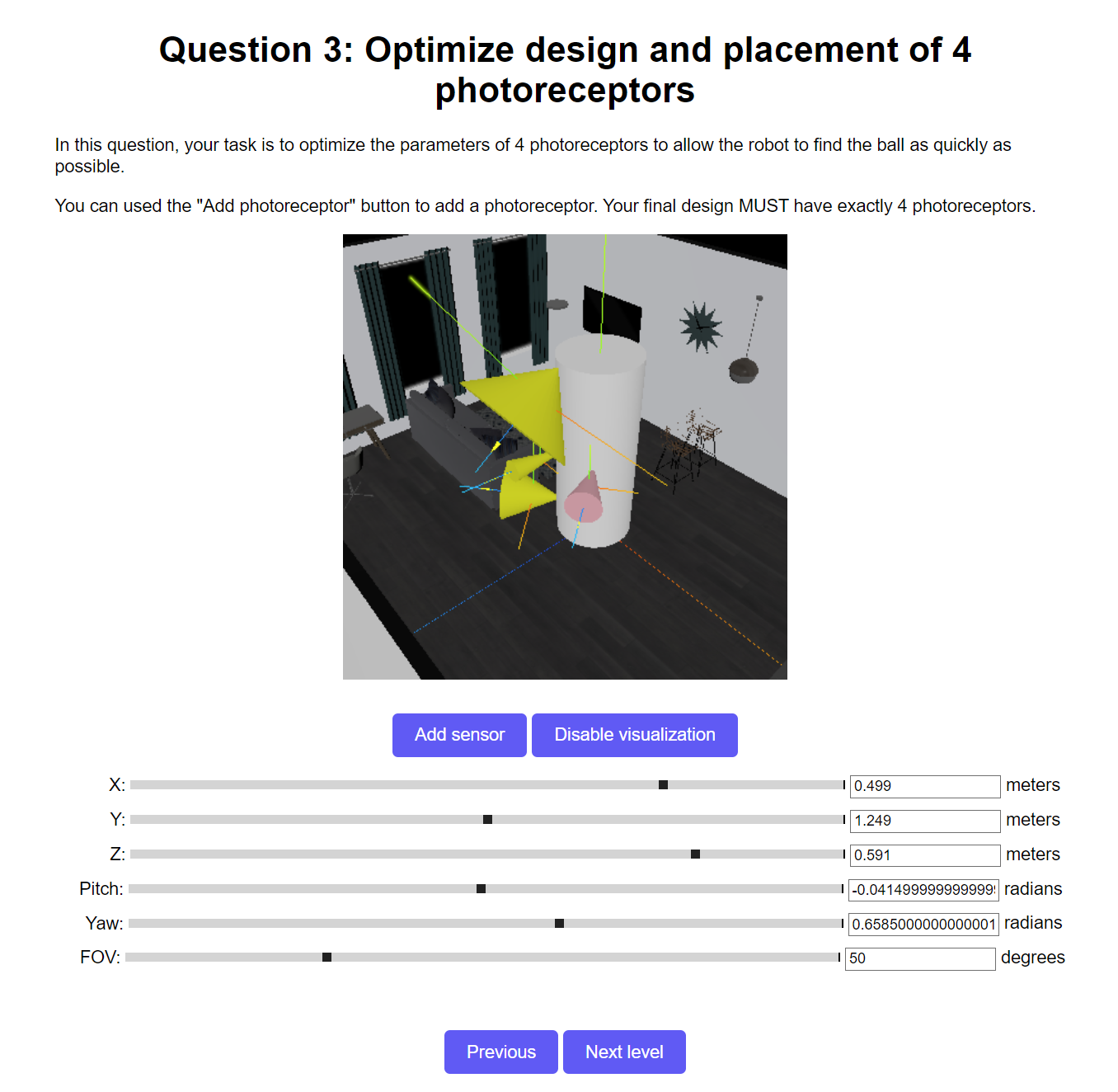}
        \caption{Level 2, question 3 (design optimization of 4 photoreceptors)}
        \label{fig:survey-lvl2}
    \end{subfigure}
    \caption{\textbf{Examples of different questions in different levels.} The cylinder is the robot body, yellow cones are sensors (either cameras or photoreceptors), and the pink cone is the selected, editable sensor. The field of view is visualized through the width of the sensor "frustum".}
    \label{fig:survey-qs}
\end{figure}

\begin{figure}
    \centering
    \includegraphics[width=0.8\linewidth]{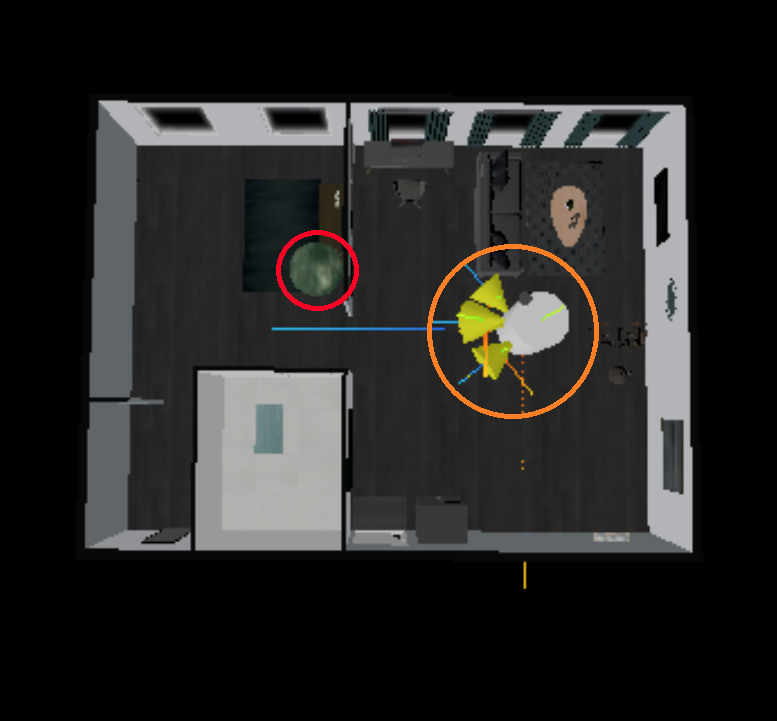}
    \caption{A bird's eye view of the environment in Level 2. The target is circled in red, and the robot is circled in orange.}
    \label{fig:survey-bev}
\end{figure}

The survey consists of three levels, each containing three questions, making a total of nine questions. Each level provides participants with progressively more specific context regarding the robot's environment. Each question focuses on optimizing the parameters of one or more visual sensors. To aid visualization, every question includes an interactive 3D render of the robot and its sensors. \cref{fig:survey-qs} illustrates examples of different questions and levels in the survey. The target across all questions and levels is a green ball with a radius of 0.5 meters, positioned 1.5 meters above the ground.
\vspace{3mm}
Questions in each level:
\begin{enumerate}[leftmargin=4em]
    \item Optimize the 3-dimensional position, pitch, yaw, and FoV of two 1 $\times$ 1 photoreceptors.
    \item Optimize the 3-dimensional position, pitch, yaw, and FoV of four 1 $\times$ 1 photoreceptors.
\end{enumerate}
\vspace{3mm}

Environment context given in each of the three levels:
\begin{enumerate}[leftmargin=4em]
\item \textit{Environment-independent context}: The target is hovering 1.5 meters above the ground in a random location. Neither the ball nor the environment is shown in the 3D visualization viewport.
\item \textit{Environment-dependent context}: In addition to the context from the previous level, the robot is now rendered with a specific example environment: a true-to-scale mesh of a home environment as well as a mesh of the target 1.5 meters above the floor. The home environment is a simplified stand-in for a Matterport3D mesh. A visualization of this level's environment is given in Fig.~\ref{fig:survey-bev}.
\item \textit{Change in photoreceptor resolution}: In addition to the context from the previous two levels, the participant is informed that the each photoreceptor's design will be used for a grid of 4 $\times$ 4 PRs instead of a single PR which is of resolution 1 $\times$ 1.. This level is``optional''; if participants believe that this change will not affect their design from previous levels, they can choose to skip it.
\end{enumerate}
\vspace{3mm}

Information about the robot given in all levels:
\begin{enumerate}[leftmargin=4em]
\item At every step, the robot can take one of four possible actions: move forward 0.25 meters, turn left 30 degrees, or turn right 30 degrees.
\item The robot is controlled by a reinforcement learning (RL) trained policy that uses the visual output from the robot's sensors to navigate to the target. The RL policy rewards the robot for navigating to the ball with the least distance traveled. The policy has memory of the robot's past actions and visual inputs through an LSTM. Familiarity with RL is not required to complete the survey.
\item The robot's forward direction is along the positive Z axis.
\item The robot has a height of 2.5 meters, which is also the maximum height for sensor placement.
\end{enumerate}
\vspace{3mm}

\subsection{Continuous Control Tasks from the DMC Suite}
The DMC benchmark uses the MuJoCo simulator~\cite{todorov_mujoco_2012}, which is challenging to deploy within a browser due to the specialized format used to define scenes and the agent's morphology. Therefore, we ask participants to sketch their placement design based on a rendered image depicting the agent's morphology and the environment. Given a sketch, we implement the design inside the simulator, show it to the participant, and update it based on their feedback until convergence, i.e., when the participant agrees that the design corresponds to their intended placement.

Due to the demanding nature of this process, we collect designs for only two tasks within one domain from six participants for continuous control tasks. Each participant provides one design for both tasks (\texttt{Walker} and \texttt{Walker}), which share the \texttt{Walker} domain. We provide a description of these two tasks similar to that in the original paper introducing the DMC benchmark~\cite{tassa_dm_control_2020}.

\end{document}